\documentclass[a4paper]{article}

\usepackage[utf8]{inputenc} 
\usepackage[T1]{fontenc}    
\usepackage{amsmath,amsfonts}
\usepackage{algorithmic}
\usepackage{algorithm}
\usepackage{array}
\usepackage{graphicx}
\usepackage{subcaption}
\usepackage{tabularray}
\usepackage{multirow}
\usepackage{adjustbox}
\usepackage{url}
\usepackage{nicefrac}
\usepackage{microtype}
\usepackage{booktabs}
\usepackage{hyperref}
\usepackage{rotating}
\usepackage{stfloats}
\usepackage{textcomp}
\usepackage{verbatim}
\usepackage{anyfontsize}
\usepackage{amsmath}
\usepackage{multicol}
\usepackage[table]{xcolor}
\usepackage{rotating}
\usepackage{setspace}
\usepackage{placeins}
\usepackage{float}

\usepackage{authblk}
\usepackage[margin=1in]{geometry}

\usepackage{listings}
\usepackage{xcolor}
\usepackage{caption}

\definecolor{codegray}{rgb}{0.5,0.5,0.5}
\definecolor{backcolor}{rgb}{0.95,0.95,0.95}

\lstdefinestyle{pythonstyle}{
    backgroundcolor=\color{backcolor},
    commentstyle=\color{codegray},
    keywordstyle=\color{blue},
    numberstyle=\tiny\color{codegray},
    stringstyle=\color{red},
    basicstyle=\ttfamily\footnotesize,
    breaklines=true,
    captionpos=b,
    keepspaces=true,
    numbers=left,
    numbersep=5pt,
    showspaces=false,
    showstringspaces=false,
    showtabs=false,
    tabsize=2,
    language=Python
}

\usepackage[style=numeric, sorting=none,maxnames=10,
    minnames=10]{biblatex}

\begin{document}


\title{Unsupervised domain transfer: Overcoming signal degradation in sleep monitoring by increasing scoring realism}



\author[ *]{Mohammad Ahangarkiasari}
\author[ ]{Andreas Tind Damgaard}
\author[ ]{Casper Haurum}
\author[*]{Kaare B. Mikkelsen}
\affil[ ]{Department of Electrical and Computer Engineering, Aarhus University, Denmark}
\affil[*]{Corresponding authors: $\{$ahangar100,mikkelsen.kaare$\}@$ece.au.dk}

\maketitle

\begin{abstract}
\textbf{Objective:}
\textcolor{black}{Investigate whether hypnogram 'realism' can be used to guide} an unsupervised method for handling arbitrary types of signal degradation in mobile sleep monitoring.

\textbf{Approach:}
Combining a pretrained, state-of-the-art ‘u-sleep’ model with a ‘discriminator’ network, we align features from a target domain with a feature space learned during pretraining. To test the approach, we distort the source domain with realistic signal degradations, to see how well the method can adapt to different types of degradation. We compare the performance of the resulting model with best-case models designed in a supervised manner for each type of transfer.

\textbf{Main Results:}
Depending on the type of distortion, we find that the unsupervised approach can increase Cohen’s kappa with as little as 0.03 and up to 0.29, and that for all transfers, the method does not decrease performance. However, the approach never quite reaches the estimated theoretical optimal performance, and when tested on a real-life domain mismatch between two sleep studies, the benefit was insignificant.

\textbf{Significance:}
‘Discriminator-guided fine tuning’ is an interesting approach to handling signal degradation for 'in the wild' sleep monitoring, with some promise. \textcolor{black}{In particular, what it says about sleep data in general is interesting.} However, more development will be necessary before using it ‘in production’.


\end{abstract}

\section{Introduction}
\label{sec:introduction}

Sleep is a fundamental aspect of human health, and accurate monitoring of sleep patterns and physiological phenomena is essential for diagnosing a wide range of medical conditions~\cite{altevogt2006sleep, sateia_international_2014}. Sleep scoring, the process of identifying and classifying distinct sleep stages, plays a central role in the screening, evaluation, and diagnosis of sleep-related disorders~\cite{phan2022automatic}. The gold standard for sleep scoring is to perform it manually by trained sleep technicians, following standardized protocols. These guidelines are outlined in the “Manual for the Scoring of Sleep and Associated Events,” published by the American Academy of Sleep Medicine (AASM)~\cite{american2007aasm}. However, due to the time consuming nature of manual sleep scoring (a single recording may take two hours to score \cite{fiorillo2019automated}), it is an on-going effort to develop good, automatic sleep scoring models \cite{phan2022automatic}.

This endeavor has led to several good sleep scoring models \cite{perslev2021u,phan2022sleeptransformer}, which never the less can still struggle when there is a mismatch between training and application domains \cite{fiorillo2023usleeps, lebiecka-johansen2025benefits, heremans2022unsupervised}. This is a general problem in the machine learning literature, and many different approaches for 'transfer learning' have been proposed \cite{olivas2009handbook,phan2020personalized}. In one of our previous studies, Lorenzen et al 2024\cite{lorenzen2024personalization}, we investigated the benefit of different methods to deal with the domain mismatch,  for the specific problem of fine tuning sleep scoring models to personal domains. There, we found very limited benefit from so-called 'unsupervised' methods, which do not rely on ground truth labels from the target domain. This was unfortunate, since the reliance on ground truth (manual) labels is an obstacle to greater adoption of automated sleep scoring.

In this study, we return to the topic of applying unsupervised domain transfer to sleep scoring, but apply it instead to the problem of adapting to various realistic signal degradations ('distortions') that may occur when recording biomedical signals 'out in the wild'. This is a very real problem, particularly for dry-contact EEG sleep monitoring, where signal quality may be highly variable \cite{mikkelsen2019accurate,leach2020protocol,kleeva2024resting,mikkelsen2025ear}.

As shown below, we find that unsupervised methods, specifically 'adversarial domain transfer' \cite{ganin2016domain} show great potential in dealing with different forms of signal degradation.

\textcolor{black}{Additionally, we find that by simply searching for the most 'realistic' hypnogram that can be made to fit the data, we also find hypnograms that are closer to the actual, correct sleep scoring. This highlights the possible use of the 'sparsity' in the hypnogram space for intelligent sleep scoring algorithms, and is the main contribution of this paper.}

\section{Hypothesis}

Most manually generated hypnograms (sleep scorings), generated by scoring polysomnography data, follow a characteristic pattern; patients usually start and end the night awake, the sleep stages are usually of a certain duration, and they tend to come in a certain order. Taken together, these restrictions define a much smaller space of 'realistic' hypnograms than the total space of 'possible' hypnograms. However, when a sleep recording is severely contaminated by recording noise, or some other issue is creating a significant domain mismatch, the result is often a less realistic hypnogram, drawing from a much larger portion of the full hypnogram space. We now hypothesize that by improving the 'realism' of a hypnogram, we may also increase the accuracy, relative to a non-polluted version of the same recording. In other words, can we disentangle the signal from noise by requiring that the resulting hypnogram should look 'convincing'? \textcolor{black}{To which extent can the 'sparsity' of correct hypnograms in the full space of possible hypnograms be leveraged for model updating?}

\begin{figure}[htbp]
    \centering
    \begin{subfigure}[t]{0.5\textwidth}
        \centering
        \includegraphics[width=\textwidth]{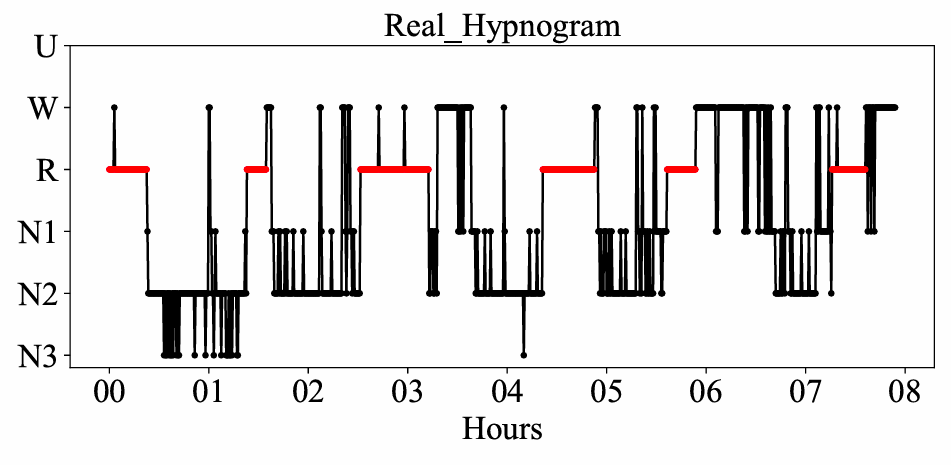}
    \end{subfigure}
    \caption{Example hypnogram illustrating the progression of sleep stages across a single night. The x-axis represents time in hours, and the y-axis denotes sleep stages based on the AASM categories: W (Wake), R (REM), N1, N2, N3, and U (Unknown/non-scorable).}
\label{fig:hypnogram_example}
\end{figure}

To explore this question, we formalize the problem of overcoming signal degradation in terms of domain adaptation. Assume clean sleep recordings as a source domain $\mathcal{D}_s = {s_1, \ldots, s_n}$, where each $s_i \in \mathbb{R}^{C \times T}$ represents a sleep recording with $C$ channels and $T$ time samples. Similarly, assume a target domain $\mathcal{D}_t = {t_1, \ldots, t_m}$, where $t_i \in \mathbb{R}^{C \times T}$ consists of recordings whose distribution is shifted relative to the source domain, due to factors such as noise, device variability, or clinical conditions.  The goal is to adapt a model trained on $\mathcal{D}_s$ to generalize sleep-stage classification on \(\mathcal{D}_t\), despite the absence of annotations in \(\mathcal{D}_t\).

To achieve this, we adopt an adversarial learning \cite{ganin2016domain} framework designed to minimize the apparent discrepancy between the two domains. Specifically, data are processed through two networks: a sleep scorer and a discriminator. The discriminator is trained to  distinguish between hypnograms derived from source and target domains \(\mathcal{D}_s\) and \(\mathcal{D}_t\). The sleep scorer aims to fool the discriminator on the target domain  while remaining accurate on the source domain. If the approach is successful, the sleep scoring network can achieve notable performance improvement on data from the target domain, without having access to any ground truth labels on which to base its adjustments.

Note that this approach does not attempt to reconstruct the original signal, or determine what the issue may have been.


\section{Related Works}

Considering specifically the topic of handling signal degradation, we can compare our work with the recent review by Raj et al \cite{raj2025comprehensive}, who looked at 29 different Deep Learning based approaches to EEG denoising, or the equally relevant works by Kalite et al \cite{kalita2024aneeg} and Gorjan et al \cite{gorjan2022removal}. What sets our approach apart from these works is that instead of trying to recreate the clean data 'underneath' the artifacts, we focus on performance on a specific task (sleep scoring), and teach a model to either ignore the artifacts or work around them. As will be seen below, this recipe allows us to deal with signal degradation using a generic solution.

As to the other aspect of our work, 'domain adaptation' aims to address the distributional mismatch between source and target data. Existing approaches, both in biomedical data and outside, typically incorporate specific objectives into the training process, such as moment matching\cite{kiasari2018joint}, Kullback–Leibler divergence \cite{phan2020personalized,hsu2017unsupervised}, or adversarial learning \cite{ganin2016domain}.

Particularly relevant for this work, Heremans et al ~\cite{heremans2022unsupervised} focused on evaluating a 'supervised' adversarial domain adaptation framework in real-world sleep staging scenarios. Borrowing the model structure originally proposed by Ganin et al. \cite{ganin2016domain}, their study explored adaptation across diverse EEG datasets, focusing on how to transfer performance from PSG models to 'wearable' models, and obtaining improvements in Cohens kappa of 0.11 to 0.22. In contrast to this approach, our model operates in a fully unsupervised setting, requiring no access to labeled data or recordings from the target domain. This design makes it particularly interesting for practical deployment scenarios where annotated data is unavailable.

Equally relevant, Yoo et al.~\cite{yoo2021transferring} investigated unsupervised domain adaptation for automatic sleep staging to address distribution shifts between labeled source datasets and unlabeled target datasets. Within an adversarial framework, they introduced local discriminators to reduce structural misalignment in sleep data. Their work differs from ours in the design of the sleep scoring model, the discriminator, the amount of data used (Yoo et al investigate transfer between two different datasets, where we aim for more mixed domains), and the fact that they transfer between non-distorted domains. This last point may partly explain why the improvements they find are generally smaller than what is found here.

Finally, Fan et al.~\cite{fan2022unsupervised} proposed an unsupervised domain adaptation approach for automatic sleep staging based on domain statistics alignment in batch normalization layers. By aligning batch-normalization statistics between source and target domains, their method reduces distribution mismatch and improves target-domain performance without requiring labeled target data. Compared to the present study, their method is not adversarial, the transfer is from a single, homogeneous source domain, and the target domains are all 'clean' datasets.

In combination, our work here both relates to existing studies in this field, but also adds a new approach, \textcolor{black}{and we are not aware of any studies investigating the 'realism' aspect.}

\section{Methods}










\subsection{Data}

In this study, we primarily use open data sets from the National Sleep Research Resource\cite{zhang2018national}, similar to what was used in Perslev et al \cite{perslev2021u} and Strøm et al~\cite{strom2024common}. Specifically, the datasets used for training and testing are given in Table \ref{tab:data_description}. This means that we are only using multichannel datasets with both EEG and EOG channels available. For these, we prioritized picking either the C4-LPA derivation or the C3-RPA as the EEG channel, and the EL-RPA or the ER-LPA as the EOG channel \cite{nuwer201810-10}. For recordings that were missing either both EEG or both EOG derivations, a random derivation of the same type was used instead.

To reduce the confounding effects of human error in our analysis, specifically that stemming from incorrect manual scorings being used as ground truth, we used the output from a pretrained U-sleep model as the ground truth labels (prior to adding noise or distortions to the data). 


As mentioned, we treat data cleaning as a form of domain transfer, between 'clean' (source) and 'distorted' (target) domains. Therefore, we start from a diverse selection of 'clean' source datasets, and then artificially add various types of 'distortions' to create target domains over which we have full control:

\begin{table}
\footnotesize
\centering
\setlength{\tabcolsep}{4pt} 
\renewcommand{\arraystretch}{1.2} 
\begin{tabular}{lcccccc}
\hline
\textbf{Dataset} & \textbf{Subjects} & \textbf{Records} & \textbf{Dataset} & \textbf{Subjects} & \textbf{Records} \\
\midrule

ABC & 49 & 132   & CCSHS     & 515 & 515 &   \\
CFS        & - & 730   & DCSM & 255 & 255  \\
DOD-H      & 25 & 25   & DOD-O     & 55 & 55  \\
EESM19     & 20 & 200   & HOMEPAP   & - & 233  \\
ISRUC-SG1  & 100 & 91  & ISRUC-SG2 & 8 & 6  \\
ISRUC-SG3  & 10 & 10   & MASS-C1   & 53 & 40 \\
MASS-C3    & 62 & 61   & PHYS      & 994 & 919 \\
SEDF-SC    & 78 & 145   & SEDF-ST   & 22 & 36 \\
SOF        & 453 & 453   & SVUH      & - & 25 \\

\bottomrule
\end{tabular}
\caption{Datasets used in this study. Further details of each data set are found in the following references:
ABC: \cite{bakker2018gastric}
CCSHS: \cite{zhang2018national},
CFS:\cite{redline1995familial},
DOD:\cite{guillot2020dreem},
EESM19: \cite{mikkelsen2025ear},
HOMEPAP: \cite{rosen2012multisite},
DCSM:\cite{zhang2018national},
 ISRUC:\cite{khalighi2016isruc},
 MASS:\cite{oreilly2014montreal},
 PHYS:\cite{ghassemi2018you},
 SEDF :\cite{kemp2000analysis},
 SVUH:\cite{goldberger2000physiobank},
 SOF:\cite{spira2008sleep}. Where possible, data was obtained through the 'National Sleep Research Resource'\cite{zhang2018national}.
}
\label{tab:data_description}
\end{table}





\textbf{Source Domain:} The datasets used in this study are summarized in Table~\ref{tab:data_description}. They comprise a diverse collection of publicly available sleep datasets, each differing in recording conditions, subject demographics, and data quality. As can be seen from the table, we have focused on getting a diverse selection of datasets.

\textbf{Target domains:}

For training and evaluation, we constructed artificially distorted datasets by introducing various types of 'distortions' to the source domain. The selected distortions were designed to challenge the sleep scoring model without rendering the task infeasible, thereby necessitating adaptation of the model to maintain performance.

Briefly, the investigated distortion types were:

\begin{description}
    \item[\textbf{White noise}] \hfill \\ Intended to simulate one of the input channels being broken, rendering the resulting channel useless. We do this by changing that channel to white noise. Since the U-Sleep model distinguishes between EEG and EOG derivations, we test distorting these separately. The standard deviation of the white noise distortion is set to 5.
    \item[\textbf{Amplifier overload}] \hfill \\ Problems with bad electrode connections and bad ground connection may result in large spikes interrupting the signal. We simulate this by adding a bi-exponential transient to sleep signals, producing a rapid overshoot followed by an equally rapid undershoot spike. In this case, the amplitude was set as twice the amplitude of the clean signal. An example is shown in the appendix. To simulate its occurrence more realistically in experimental settings, white noise with standard deviation 1 is superimposed onto this pulse. As shown in the results section, we test different frequencies of occurrence for this distortion.
    \item[\textbf{Spectral deformation}] \hfill \\ Both hardware and physiology may result in altered distributions of power in the frequency domain. We simulate this by running both signals through a 4th-order digital IIR band-pass filter, implemented in a numerically stable SOS form, to keep  frequencies between \textit{lowcut} and \textit{highcut} and reduce everything else. We fix \textit{highcut} at 20 Hz, and test different values of \textit{lowcut}.
\end{description}

See the appendix for visualizations.


\subsection{Model implementation}
\label{Model implementation}

\begin{figure*}[t]
    \centering
    \begin{subfigure}[t]{0.9\textwidth}
        \centering
        \includegraphics[width=\textwidth]{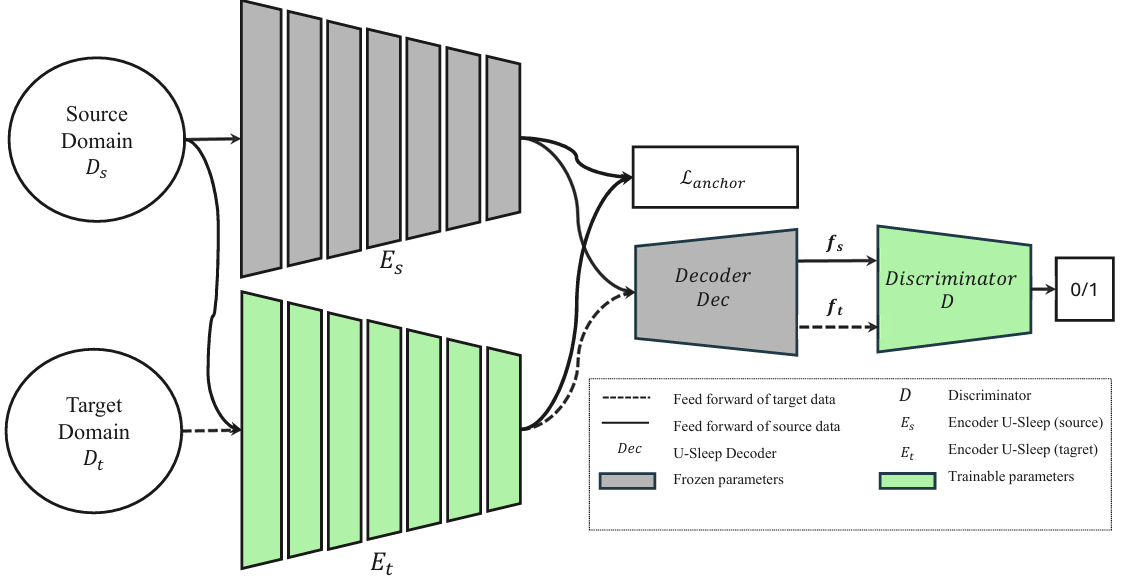}
    \end{subfigure}
    \caption{The overall structure of the proposed model. \( E_{\text{s}} \) denotes the frozen, pretrained encoder used for the source domain, while \( E_{\text{t}} \) is initialized with the same pretrained U-Sleep weights but is fine-tuned during training to adapt to the target domain. \( Dec \) is the pretrained U-Sleep Decoder and \(D\) is the discriminator. $f_s$ and $f_t$ denote the decoder output features of the U-Sleep model for the source and target domains, respectively (refer to section~\ref{Model training}).}
\label{fig:model}
\end{figure*}

We focus on a specific implementation of our framework, illustrated in Figure \ref{fig:model}:

The full model consists of a sleep scoring network and discriminator. In our design, we use the Usleep model by Perslev et al \cite{perslev2021u} as the basis for the sleep scoring network. To enforce the hypothesis being tested, namely that the model should learn to 'work around' the distortion to still generate plausible sleep scorings, we split Usleep in its encoder and decoder portions, and will treat these slightly differently:

\subsubsection{Encoder}
Copying the original U-sleep implementation, the encoder comprises $12$ sequential encoder blocks. Each block includes a convolutional layer followed by an exponential linear unit (ELU) activation function, batch normalization, and max-pooling. The number of filters increases progressively with each block, starting from five filters in the initial layer and growing by a fixed factor in subsequent layers. The encoder’s main role is to extract increasingly abstract feature maps from the input signals while simultaneously reducing their temporal resolution, enabling the model to capture essential patterns for accurate sleep stage classification. In our training framework, two encoders are employed: one for the source domain $E_s$, which is frozen and provides clean features, and another for the target domain, $E_t$ which extracts distorted features from the target domain. Both encoders are initialized with weights from a pretrained Usleep.

\subsubsection{Decoder}
The decoder module also consists of $12$ decoder blocks, each performing upsampling using nearest neighbor interpolation followed by a convolutional layer, ELU activation, and batch normalization. After upsampling, the output is merged through skip connections with the corresponding encoder block outputs taken before max-pooling. These skip connections link layers that share the same temporal resolution; for example, the last decoder block aligns with the first encoder block. Unlike the encoder, the decoder reduces the number of learned filters by a factor of two at each successive layer to mirror the encoder’s temporal resolution changes. Once processed through the decoder, the output represents a high-frequency, detailed sleep stage signal that aligns with the input sampling rate of 128 Hz. Our encoder also incorporates the 'segment classifier', which in the original U-Sleep architecture is implemented as a separate block. The segment classifier aggregates the high-frequency feature representations generated by the decoder over longer time segments to produce sleep stage predictions. Since the goal is to generate one prediction per 30-second epoch, the segment size is set accordingly. Unlike the encoder, the decoder is 'frozen', meaning the weights are not updated. 

\subsubsection{Discriminator}

The discriminator is implemented as a lightweight Transformer-based binary classifier. Initially, the input is fed into a linear projection layer that maps each channel to a fixed embedding dimension, to which is added a learnable positional embedding to preserve temporal order. The result is processed through a stack of 2 Transformer encoders composed of multi-head self-attention (4 heads) and feedforward sub-modules. After the encoders, the sequence is flattened and passed through a fully connected layer to produce a scalar output. A sigmoid activation function is applied to indicate whether the input originated from the source or target domain. Note that the discriminator receives the predicted class probability distributions across all five sleep stages. In this way, the discriminator learns to distinguish between the probabilistic output patterns of the source and target domains. Thus, it can be interpreted as learning the distributional pattern of predicted sleep-stage probabilities, rather than making decisions based on hard class assignments. In our implementation, we instantiate three such discriminators in parallel to form the final domain prediction. This is found to provide more reliable adversarial supervision in our experiments. The structure of the model is fixed throughout all experiments.



\subsection{Model training}
\label{Model training}
The overall training process is formulated through two components: an anchor loss and an adversarial loss. These objectives jointly guide the model to learn domain-invariant and semantically consistent representations during adaptation.

To support these objectives, the source and target encoders $E_s$ and $E_t$, and the decoder $Dec$ are initialized from a pretrained U\textsc{-}Sleep, while the discriminator $D$ is randomly initialized. Let $f_s = Dec(E_s(r_s))$ and $f_t = Dec(E_t(r_t))$ denote the source and target features that are fed to $D$, where $r_s \sim \mathcal{D}_s$ and $r_t \sim \mathcal{D}_t$ (denote samples drawn from the \emph{empirical} source and target data distributions induced by the datasets $\mathcal{D}_s$ and $\mathcal{D}_t$, respectively). The expectations $\mathbb{E}_{r_s \sim \mathcal{D}_s}[\cdot]$ and $\mathbb{E}_{r_t \sim \mathcal{D}_t}[\cdot]$ are approximated by mini-batch averages during training. The discriminator is trained with a binary cross-entropy objective to output $0$ for target features and $1$ for source features:
\begin{equation}
\mathcal{L}_{\text{adv}}^{D}
= \,\mathbb{E}_{r_s \sim \mathcal{D}_s}\!\left[\log \big(1 - D(f_s)\big)\right]
\;+\; \mathbb{E}_{r_t \sim \mathcal{D}_t}\!\left[\log \big( D(f_t)\big)\right],
\label{eq:LD}
\end{equation}
so minimizing $\mathcal{L}_{\text{adv}}^{D}$ pushes $D(f_s)\!\to\!1$ and $D(f_t)\!\to\!0$.  In contrast, the target encoder is optimized to \emph{fool} the discriminator, i.e., to make target features appear source-like. To train $E_t$, we therefore maximize the discriminator outputs on target features, which is implemented by Eq.~\ref{eq:LE}, driving $D(f_t)\!\to\!1$.
\begin{equation}
\mathcal{L}_{\text{adv}}^{E_t}
= \,\mathbb{E}_{r_t \sim \mathcal{D}_t}\!\left[\log (1-D(f_t))\right],
\label{eq:LE}
\end{equation}
To stabilize adaptation and preserve the pretrained geometry on source data, we further introduce an anchor loss that penalizes deviations between the target and source encoders on the \emph{same} source inputs:
\begin{equation}
\mathcal{L}_{\text{anchor}}
= \mathbb{E}_{r_s \sim \mathcal{D}_s}\!\left[\left\|E_t(r_s) - E_s(r_s)\right\|_2^{2}\right].
\label{eq:LA}
\end{equation}

At each iteration, we update $D$ by minimizing $\mathcal{L}_{\text{adv}}^{D}$ and update $E_t$ (and any shared parameters with $Dec$ if applicable) by minimizing
\[
\alpha\,\mathcal{L}_{\text{adv}}^{E_t} + \beta\,\mathcal{L}_{\text{anchor}},
\quad \text{with } \alpha=1 \text{ and } \beta=0.1.
\]
This min–max interplay aligns the target feature distribution with the source while the anchor term enforces cross-domain consistency.

The values of $\alpha$ and $\beta$ were set apriori based on experience from previous projects, and were not tuned during these experiments.

It may be worth noting that the implementation of the anchor loss forces the model to assume 'covariate shift', meaning that if the target domain is a mixture of distorted and clean data (which is a likely scenario), the non-distorted data will still be correctly scored.

\textcolor{black}{See the appendix for plots of model losses and performance measures as a function of training epochs.}

\subsubsection{Technical implementation}

The model was trained with a batch size of 1 (full recording). \textcolor{black}{This rather small batch size was chosen to allow stable training on a laptop, even while experimenting with different model implementations. We have since determined that the present implementation does not appreciably benefit from increasing the batch size up to 8. A benefit of using single recordings is that we have not needed to adjust the data loading to accommodate varying recording lengths (by padding short recordings, for instance). }

The learning rate was set to $1 \times 10^{-5}$, for the encoder and  $1 \times 10^{-6}$ for the discriminator. Keeping the encoder’s learning rate one order of magnitude higher than that of the discriminator allows the generator to continuously adapt to the discriminator’s feedback, thereby stabilizing the adversarial training process. All parameters were optimized using the Adam optimizer, and the model was trained for 50 epochs. \textcolor{black}{A single run took between 6 and 8 hours on a laptop with dedicated  GPU. }

\subsection{Performance Evaluation}

We evaluated the performance of the pretrained and fine-tuned U-Sleep models using some complementary analyzes designed to capture both predictive accuracy and model behavior under signal distortions.

\subsubsection{Cohen’s kappa}

To quantify the agreement between predicted and true hypnograms, we use Cohen’s kappa score~\cite{cohen1960coefficient}, hereafter denoted simply as \textit{kappa}. This metric provides a standard measure of reliability across sleep stages, accounting for the level of agreement that could be expected by chance. In addition, we perform a \textit{chance kappa} analysis to verify that improvements in agreement are not attributable to generic or signal-independent predictions. 

\subsubsection{Chance kappa analysis}
To ensure that the observed improvements in kappa are not merely due to producing generic hypnograms that resemble typical sleep stage distributions without basing it on the input signal, we perform a Chance kappa analysis. This involves estimating the distribution of kappa scores obtained under random hypnogram sampling, providing a baseline against which to assess whether the fine-tuned model generates outputs that more closely follow the true temporal structure of sleep stages than the pretrained model, particularly on noisy or distorted target-domain data. We estimate the chance kappa distribution by randomly selecting two sleep recordings from the source domain and processing them with the pretrained U-Sleep model to generate two hypnograms. The agreement between these hypnograms is quantified using Cohen’s kappa score~\cite{cohen1960coefficient}. This procedure is repeated to obtain the Chance Kappa distribution for the source domain using the pretrained model. A similar process is applied to the target domain using the fine-tuned U-Sleep model. The resulting distributions are analyzed to ensure that the fine-tuned model’s outputs reflect the inherent variability of the input signals, rather than showing artificially high agreement or converging to repetitive patterns. \textcolor{black}{See the appendices for a pseudo-code description of the algorithm.}



\subsubsection{'Benchmark performance'}

To have a point of reference for each type of distortion, we compare with a benchmark model that is likely to be the best possible performance achievable (a 'native' model). For the white noise distortion, this is simply a model that ignores the white noise channel (by setting the input weights to zero, mimicing the empty channels used in pretraining), while for the 'amplifier overload' and 'spectral deformation' distortions we compare with a model trained in a conventional, supervised fashion using target labels together with distorted data.

\subsubsection{Permutation tests}

Statistical significance was assessed using a paired permutation test on per-record performance differences. For each record \textit{i}, we calculated the difference $d_{i} = m_{i}^{fine} - m_{i}^{pre}$, where $m_{i}^{fine}$ and $m_{i}^{pre}$ denote the metric of interest (F1 scores, Accuracy, Cohen’s kappa scores) based on fine-tuned and pre-trained models, respectively. The observed mean difference $\Delta_{\mathrm{obs}} = \frac{1}{N}\sum_{i=1}^{N} d_i$ was compared to a null distribution generated by randomly flipping the sign of each $d_i$ with probability of 0.5 by multiplying it with an independent random sign \(s_i^{(b)} \in \{+1,-1\}\), where \(\Pr(s_i^{(b)}=+1)=\Pr(s_i^{(b)}=-1)=0.5\). Thus, \(d_i^{(b)} = s_i^{(b)} d_i\). The permuted mean difference is then calculated as $\Delta_{\mathrm{perm}}^{(b)} = \frac{1}{N}\sum_{i=1}^{N} d_i^{(b)}$, repeating this procedure over $P = 10,000$ permutations. We then estimate the p-value as the following equation;

\begin{equation}
p = \frac{1 + \sum_{b=1}^{P} \mathbb{I}\!\left(\Delta_{\mathrm{perm}}^{(b)} \ge \Delta_{\mathrm{obs}}\right)}{1 + P},
\label{eq:pvalue}
\end{equation}

where \(  \sum_{b=1}^{P} \mathbb{I}\!\left(\Delta_{\mathrm{perm}}^{(b)} \ge \Delta_{\mathrm{obs}}\right) \) count the number of times that the \(\Delta_{\mathrm{perm}}^{(b)} \) is equal or bigger that than \(\Delta_{\mathrm{obs}}\).



\section{Results}

\subsection{Chance Kappa Evaluation of Fine-Tuned U-Sleep Model}

\begin{figure}[h]
    \centering
    \begin{subfigure}[t]{0.48\textwidth}
        \centering
        \includegraphics[width=\textwidth]{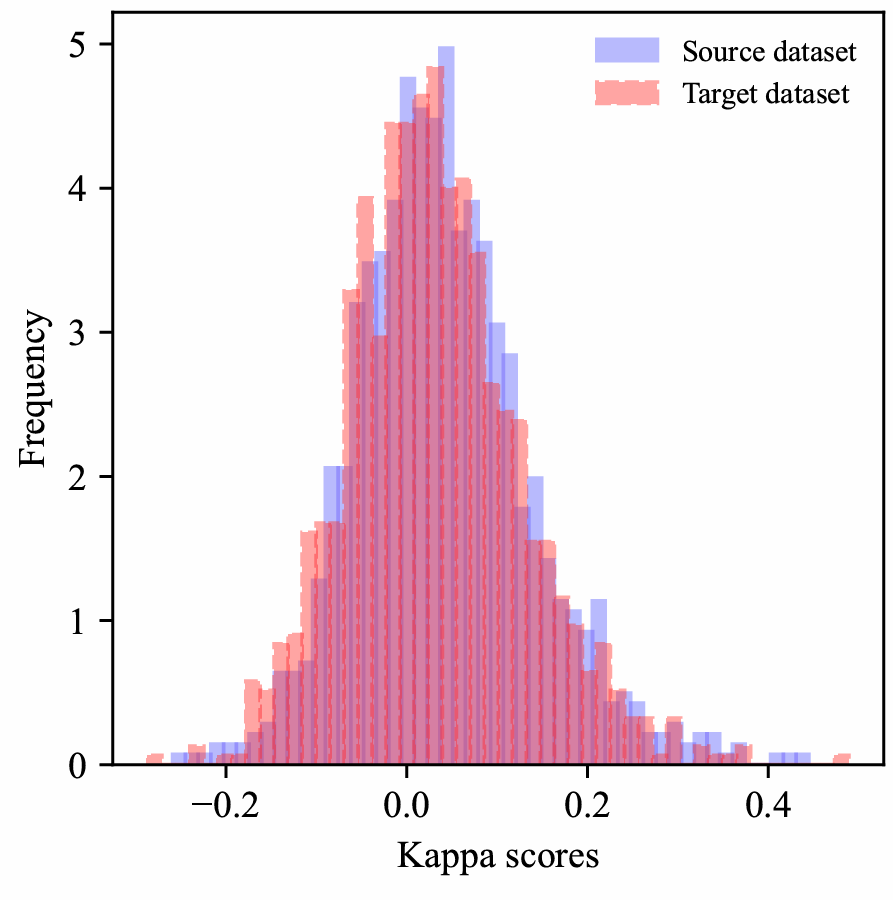}
    \end{subfigure}
\caption{Histogram of Chance Kappa scores. Kappa scores are computed from random EEG-EOG sample pairs drawn from the target and source datasets. Blue: Pretrained U-Sleep on the source dataset. Red: Fine-tuned U-Sleep on the target dataset.}
\label{fig:chance_kappa_histogram}
\end{figure}

Figure~\ref{fig:chance_kappa_histogram} presents the histograms obtained from the Chance Kappa analysis for both the source and the target domains. For each domain, 300 samples were randomly selected to compute the Chance Kappa distribution. The Chance Kappa distributions are centered around low kappa values, reflecting the variability in hypnogram predictions when PSG recordings are randomly paired (unrelated PSG recordings).  The shape and location of the two distributions are similar, suggesting that fine-tuning the U-Sleep model on the target domain has not led to output collapse or artificially high agreement across predictions. This indicates that the model continues to generate distinct hypnograms for different PSG samples, preserving the diversity of its predictions. The low mean kappa values ( \(\approx 0\) ) in both distributions confirm that agreement due to distribution bias remains low.

We see that kappa values above 0.35 can safely be considered 'significant' agreement.



\subsection{Distortion Analysis}
As described above, we test how the approach handles different types of distortions:
A total of 3,931 recordings compiled from all available datasets are utilized for this analysis. A stratified random split is performed, allocating 80\% of the data for training and the remaining 20\% for testing.

\subsubsection{White Noise}
\begin{figure*}[htbp]
    \centering
    \begin{subfigure}[t]{0.24\textwidth}
        \centering
        \includegraphics[width=\textwidth]{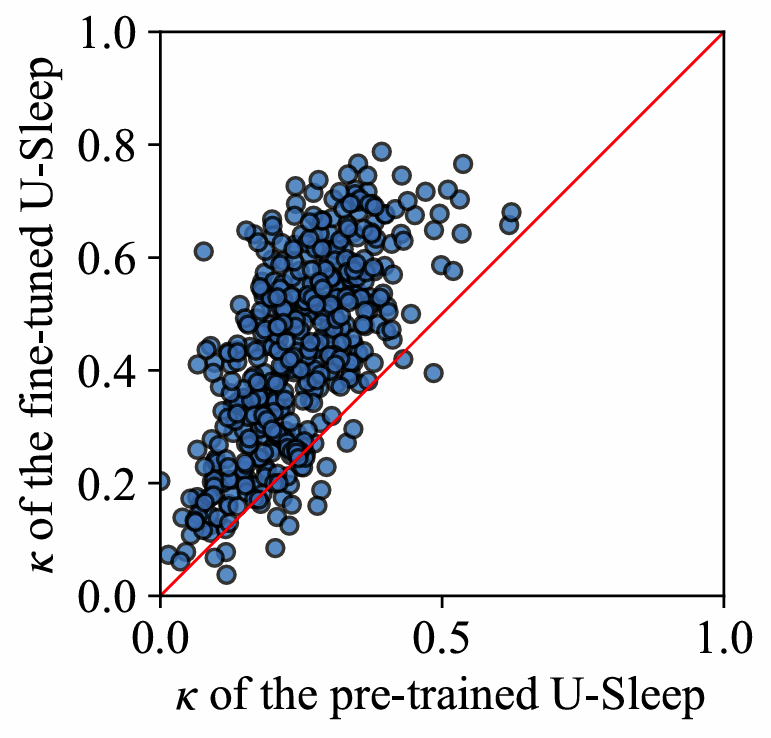}
        \caption{Finetuned vs. pretrained, distorting EEG channel.}
    \end{subfigure}
    \begin{subfigure}[t]{0.24\textwidth}
        \centering
        \includegraphics[width=\textwidth]{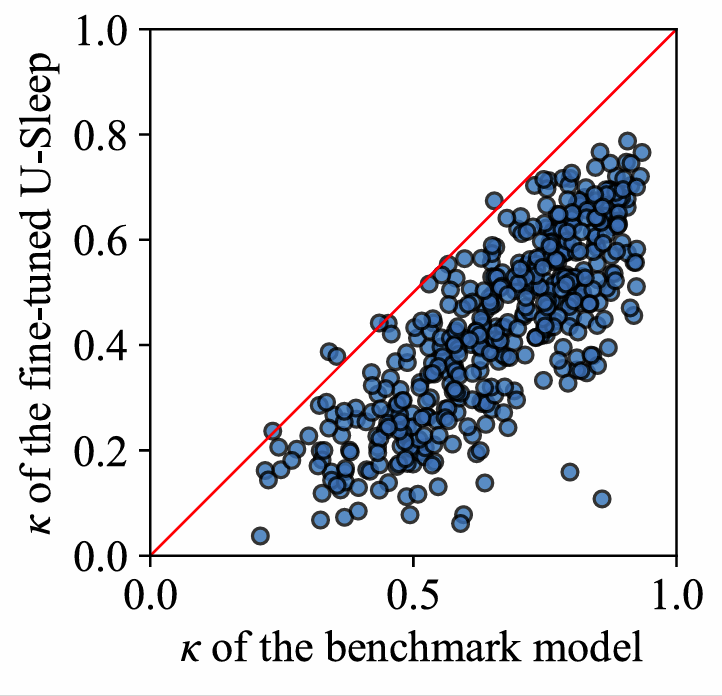}
        \caption{Finetuned vs. benchmark model, distorting EEG channel.}
    \end{subfigure}
        \begin{subfigure}[t]{0.24\textwidth}
        \centering
        \includegraphics[width=\textwidth]{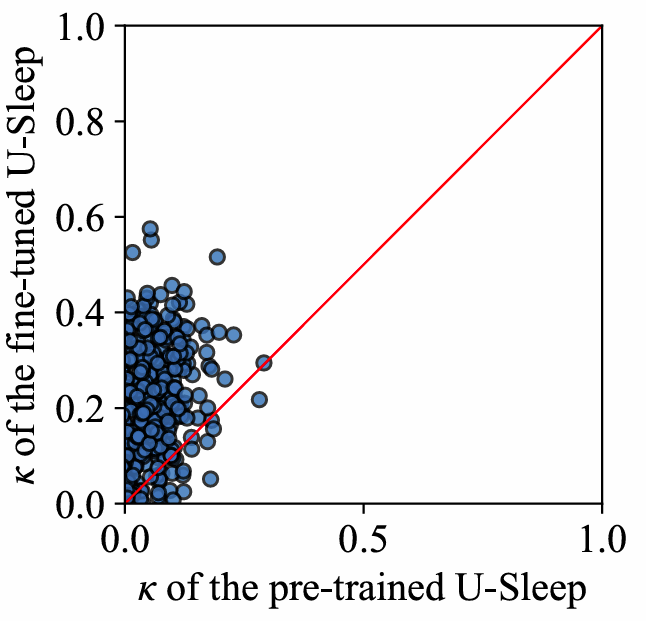}
        \caption{Finetuned vs. pretrained, distorting EOG channel.}
    \end{subfigure}
    \begin{subfigure}[t]{0.24\textwidth}
        \centering
        \includegraphics[width=\textwidth]{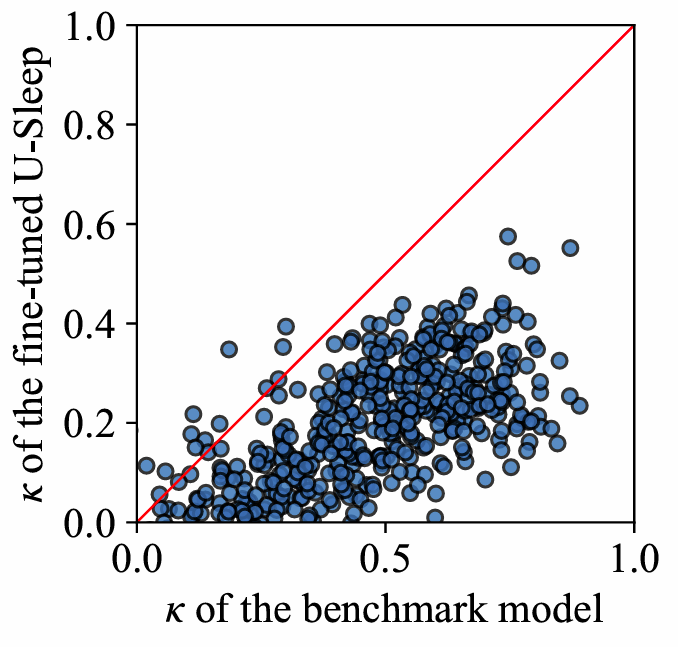}
        \caption{Finetuned vs. benchmark model, distorting EOG channel.}
    \end{subfigure}
\caption{Comparison of Cohen’s kappa ($\kappa$) before and after fine-tuning with  white noise distortion. Results are compared to both pretrained model and benchmark model.}
\label{fig:exp1}
\end{figure*}

Figure \ref{fig:exp1} presents a comparison of kappa scores between the pretrained, the fine-tuned and benchmark models, while Table \ref{tab:white_noise_results_all} reports the corresponding average values for Cohen’s kappa scores, Accuracy and F1-score. Across the board, finetuning gives a marked improvement over pretraining, but is still beaten  by the benchmark model.

We see slightly different behavior depending on which channel is distorted, reflecting differences in the model's reliance on the EEG and EOG inputs.

\begin{table}[htbp]
\footnotesize

\centering
\setlength{\tabcolsep}{6pt}
\renewcommand{\arraystretch}{1.2}
\begin{tabular}{llcc}
\hline
\textbf{Metric} & \textbf{Model} & \textbf{EEG} & \textbf{EOG} \\
\hline

\multirow{3}{*}{$\kappa$}
 & Pretrained model & 0.23 & 0.02 \\
 & Finetuned model  & \textbf{0.42} & \textbf{0.18} \\
 & $\Delta$ (Fine--Pre) & 0.29 & 0.16 \\
& \textit{p-value} & { $<1\times10^{-4}$}  & { $<1\times10^{-4}$} \\
\hline

\multirow{3}{*}{Accuracy}
 & Pretrained model & 0.46 & 0.31 \\
 & Finetuned model  & \textbf{0.60} & \textbf{0.42} \\
 & $\Delta$ (Fine--Pre) & 0.14 & 0.11 \\
 & \textit{p-value} & { $<1\times10^{-4}$}  & { $<1\times10^{-4}$} \\
\hline

\multirow{3}{*}{F1-score}
 & Pretrained model & 0.47 & 0.32 \\
 & Finetuned model  & \textbf{0.59} & \textbf{0.41} \\
 & $\Delta$ (Fine--Pre) & 0.12 & 0.09 \\
 & \textit{p-value} & { $<1\times10^{-4}$}  & { $<1\times10^{-4}$} \\
\hline
\end{tabular}
\caption{Effect of finetuning to ignore a noisy input channel. The column header indicates which U-Sleep input was replaced with white noise. Results are reported for Cohen’s $\kappa$, accuracy, and F1-score.  P-values test whether pretrained and fine-tuned performance might be the same. }
\label{tab:white_noise_results_all}
\end{table}


\subsubsection{Amplifier Overload }

Table~\ref{table:table_amplifier_overload_all} reports the model performance across amplifier-overload distortion frequencies, comparing kappa scores for pre-trained and fine-tuned models with distortion applied equally to both EEG and EOG channels. The amplifier overload frequencies are set to 20, 25, and 30 Hz. Across amplifier-overload frequencies, the fine-tuned model improves performance in terms of Cohen’s kappa, accuracy, and F1 score compared to the pre-trained baseline; however, the performance of both models degrades as the distortion frequency increases.

As with the white noise distortion, our unsupervised fine tuning does not compete with the benchmark model, finetuned with labels.



\begin{table}[t]
\footnotesize

\centering
\setlength{\tabcolsep}{1pt}
\renewcommand{\arraystretch}{1.2}
\begin{tabular}{llccc}
\hline
& & \multicolumn{3}{c}{\textbf{Frequency (Hz)}} \\
\cline{3-5}
\textbf{Metric} & \textbf{Model} & \textbf{20} & \textbf{25} & \textbf{30} \\
\hline

\multirow{3}{*}{$\kappa$}
& Pretrained  & 0.30 & 0.29 & 0.28  \\
& Fine-tuned  & \textbf{0.41} & \textbf{0.38} & \textbf{0.36}  \\
& $\Delta$ (Fine--Pre) & 0.11 & 0.09 & 0.08 \\
& \textit{p-value} & {{ $<1\times10^{-4}$}} & {{ $<1\times10^{-4}$}} & {{ $<1\times10^{-4}$}} \\

\hline

\multirow{3}{*}{Accuracy}
 & Pretrained  & 0.49 & 0.48 & 0.48  \\
 & Fine-tuned  & \textbf{0.58} & \textbf{0.55} & \textbf{0.53}  \\
 & $\Delta$ (Fine--Pre) & 0.09 & 0.06 & 0.05 \\
 & \textit{p-value} & {{ $<1\times10^{-4}$}} & {{ $<1\times10^{-4}$}} & {{ $<1\times10^{-4}$}} \\
\hline

\multirow{3}{*}{F1-score}
 & Pretrained  & 0.50 & 0.50 & 0.50 \\
 & Fine-tuned  & \textbf{0.54} & \textbf{0.54} & \textbf{0.53}  \\
 & $\Delta$ (Fine--Pre) & 0.04 & 0.05 &  0.03 \\
 & \textit{p-value} &{ $<1\times10^{-4}$} &{ $<1\times10^{-4}$} &{ $<1\times10^{-4}$} \\
\hline

\end{tabular}
\caption{Effect of amplifier-overload distortion frequency on performance. Results are reported for Cohen’s $\kappa$, Accuracy, and F1-score. The fine-tuned model exceeds the pretrained baseline across all frequencies.}
\label{table:table_amplifier_overload_all}
\end{table}

\subsubsection{Spectral deformation}

Table \ref{table:table_bad_filter_all} provides average kappa scores, accuracy, and F1-scores for different low-cut frequencies. As expected, more aggressive high-pass filtering (>5 Hz) disproportionately harms the baseline model. Across all cutoffs, the fine-tuned model improves the baseline with absolute gains $\Delta\kappa$ ={0.03,0.16,0.15,0.15} at {1,5,7,10} Hz.

\begin{figure*}[t]
    \centering
    \begin{subfigure}[t]{0.24\textwidth}
        \centering
        \includegraphics[width=\textwidth]{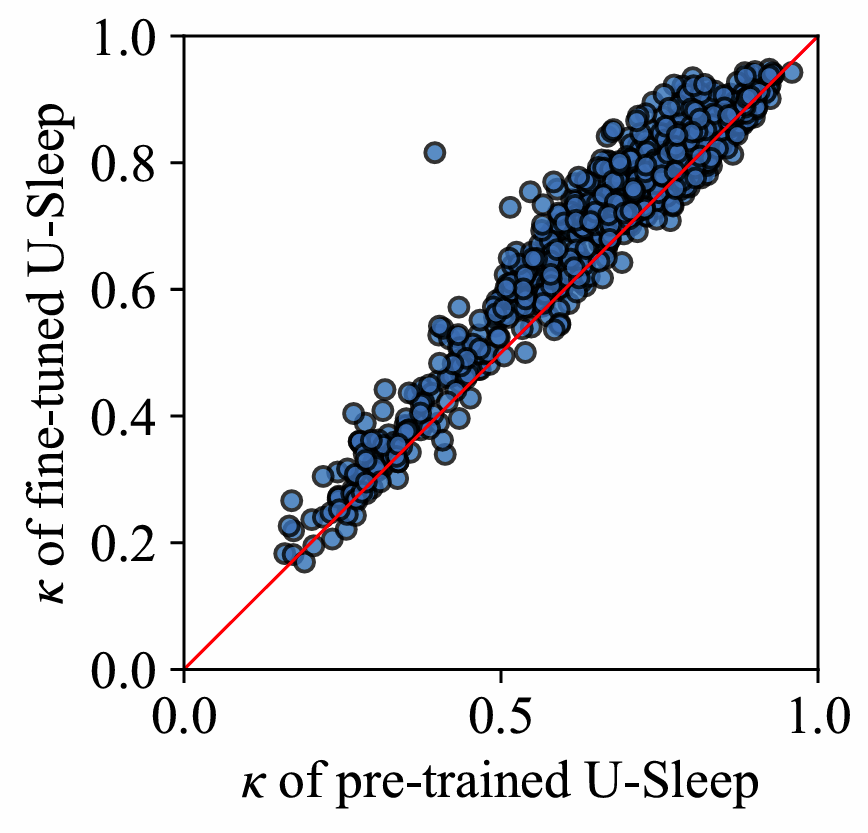}
        \caption{Low-cut: 1 Hz}
    \end{subfigure}
    \begin{subfigure}[t]{0.24\textwidth}
        \centering
        \includegraphics[width=\textwidth]{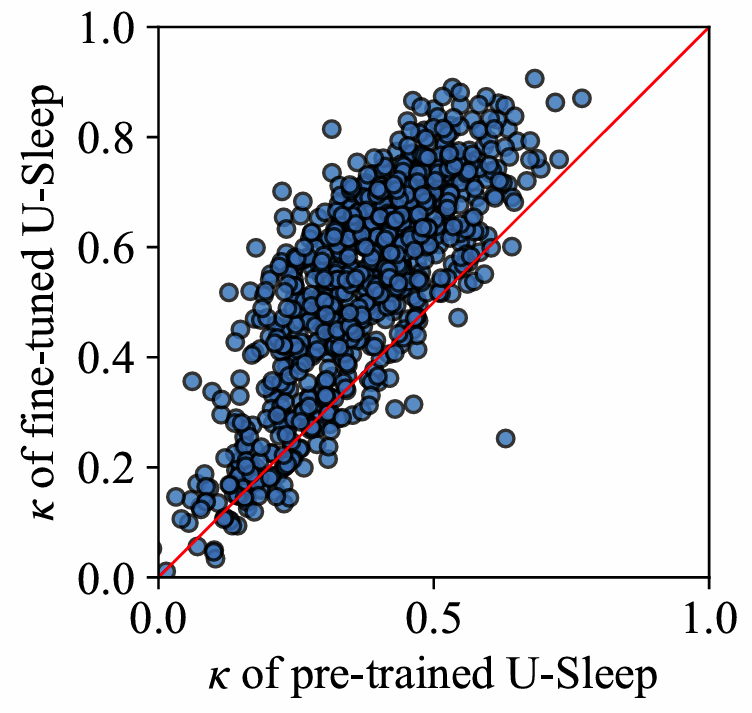}
        \caption{Low-cut: 5 Hz}
    \end{subfigure}
    \begin{subfigure}[t]{0.24\textwidth}
        \centering
        \includegraphics[width=\textwidth]{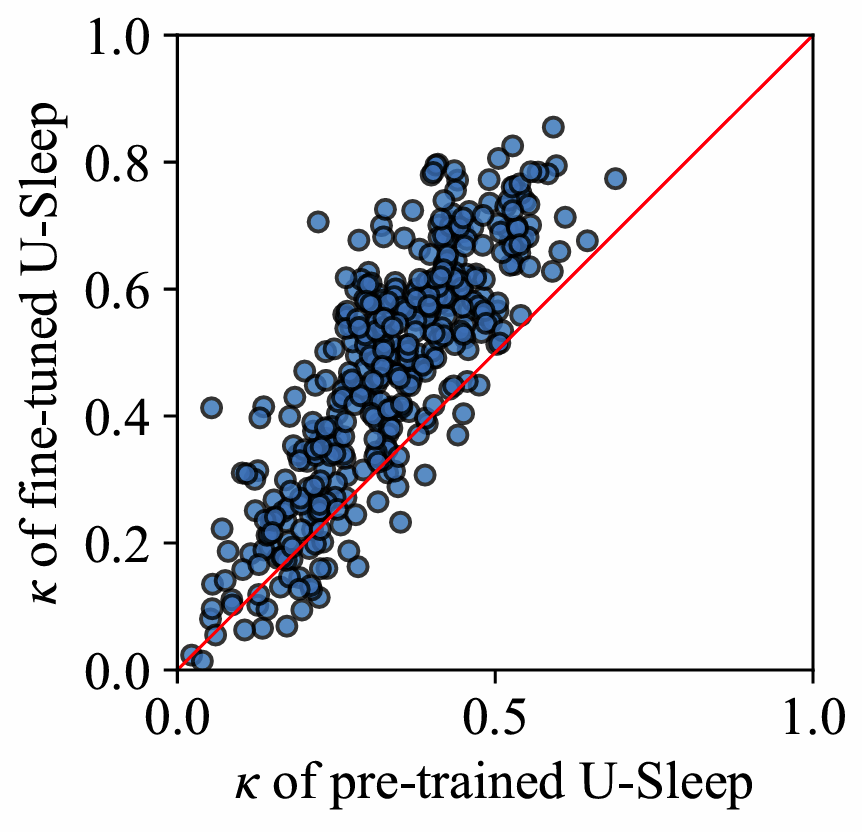}
        \caption{Low-cut: 7 Hz}
    \end{subfigure}
    \begin{subfigure}[t]{0.24\textwidth}
        \centering
        \includegraphics[width=\textwidth]{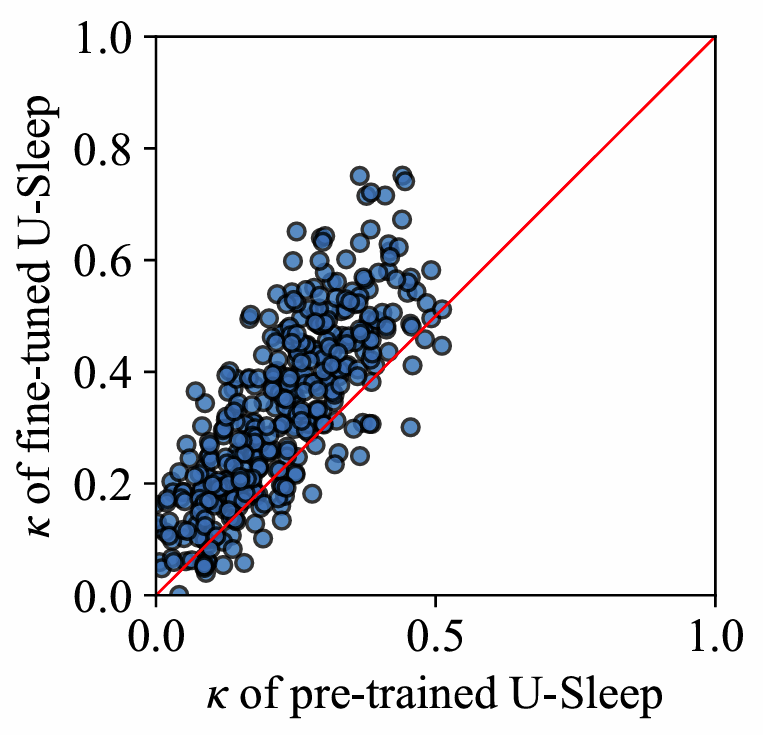}
        \caption{Low-cut: 10 Hz}
    \end{subfigure}
\caption{Cohen’s kappa ($\kappa$) between fine-tuned and pretrained U-Sleep models under different low-cut thresholds (1, 5, 7, and 10 Hz) in a mid-pass filter with a fixed high-cut frequency of 20 Hz.}
\label{fig:bad_filter_results}
\end{figure*}

A more detailed picture for kappa scores is shown in Figure \ref{fig:bad_filter_results}, where scatterplots comparing pretrained and finetuned performance are shown.

Finally, just like with the other two distortion types, we find that even though the fine tuning siginificantly improves upon the baseline performance, it is still beaten by the benchmark, 'supervisedly' trained model. We have included a scatter plot comparison in the appendix.

\begin{table*}[htb]
\footnotesize

\centering
\setlength{\tabcolsep}{10pt}
\renewcommand{\arraystretch}{1.2}
\begin{tabular}{llcccc}
\hline
& & \multicolumn{4}{c}{\textbf{Low-cut frequency (Hz)}} \\
\cline{3-6}
\textbf{Metric} & \textbf{Model} & \textbf{1} & \textbf{5} & \textbf{7} & \textbf{10} \\
\hline

\multirow{3}{*}{$\kappa$}
 & Pretrained  & 0.68 & 0.36 & 0.33 & 0.27 \\
 & Fine-tuned  & \textbf{0.71} & \textbf{0.52} & \textbf{0.47} & \textbf{0.42} \\
 & \textit{$\Delta\kappa$ (Fine–Pre)} & 0.03 & 0.16 & 0.15 & 0.15 \\
 & \textit{p-value} & { $<1\times10^{-4}$} & { $<1\times10^{-4}$} & { $<1\times10^{-4}$} & { $<1\times10^{-4}$} \\
\hline

\multirow{3}{*}{Accuracy}
 & Pretrained  & 0.77 & 0.55 & 0.53 & 0.45 \\
 & Fine-tuned  & \textbf{0.79} & \textbf{0.67} & \textbf{0.63} & \textbf{0.51} \\
 & \textit{$\Delta$Acc (Fine–Pre)} & 0.02 & 0.12 & 0.10 & 0.06 \\
& \textit{p-value} & { $<1\times10^{-4}$} & { $<1\times10^{-4}$} & { $<1\times10^{-4}$} & { $<1\times10^{-4}$} \\
\hline

\multirow{3}{*}{F1-score}
 & Pretrained  & 0.76 & 0.57 & 0.55 & 0.45 \\
 & Fine-tuned  & \textbf{0.79} & \textbf{0.66} & \textbf{0.62} & \textbf{0.51} \\
 & \textit{$\Delta$F1 (Fine–Pre)} & 0.03 & 0.09 & 0.07 & 0.06 \\
& \textit{p-value} & { $<1\times10^{-4}$} & { $<1\times10^{-4}$} & { $<1\times10^{-4}$} & { $<1\times10^{-4}$} \\
\hline

\end{tabular}
\caption{Effect of increasing high-pass (low-cut) frequency on performance for different metrics. The fine-tuned model outperforms the pretrained baseline across all settings. P-values test whether pretrained and fine-tuned performance might be the same. }
\label{table:table_bad_filter_all}
\end{table*}

\subsection{Stage-wise performance}
\textcolor{black}{For all distortion types, we see that the approach improves performance for all stages. As an example, in Table \ref{tab:stage-wise} stage-wise performance metrics are shown for the ‘spectral deformation’ case.}

\begin{table}[htb]
\centering
\caption{Per-class performance comparison against GT labels (support = number of ground-truth samples(epochs) per class).}\label{tab:stage-wise}
\setlength{\tabcolsep}{6pt}
\begin{tabular}{lccccccc}
\hline
\multirow{2}{*}{Class} & \multirow{2}{*}{Support} &
\multicolumn{3}{c}{Pretrained Model} & \multicolumn{3}{c}{pre-trained Model} \\
\cline{3-5} \cline{6-8}
& & Precision & Recall & F1-score & Precision & Recall & F1-score \\
\hline
W   & 27467 & 0.5593 & 0.5485 & 0.5538 & \textbf{0.5940} & \textbf{0.6161} & \textbf{0.6048} \\
\textbf{}N1  &  6664 & 0.2412 & 0.2758 & 0.2574 & \textbf{0.2420} & \textbf{0.2938} & \textbf{0.2654} \\
N2  & 45262 & 0.5674 & 0.6449 & 0.6037 & \textbf{0.6203} & \textbf{0.6547} & \textbf{0.6370} \\
N3  & 16756 & 0.6886 & 0.2730 & 0.3910 & \textbf{0.7630} & \textbf{0.4530} & \textbf{0.5685} \\
REM & 14221 & 0.3508 & 0.4371 & 0.3892 & \textbf{0.4245} & \textbf{0.4796} & \textbf{0.4504} \\
\hline
\end{tabular}
\end{table}

\subsection{Performance under data scaling}

To assess the influence of training data quantity on model performance, we conducted a series of experiments using different proportions of the available training records. By progressively increasing the dataset size, we aim to evaluate how the model’s learning capacity and generalization ability evolve with the amount of training data. This analysis provides insight into the model’s data efficiency and highlights whether additional data could further enhance performance or if the model has reached a saturation point.

\begin{figure}[htb]
    \centering
    \begin{subfigure}[t]{0.48\textwidth}
        \centering
        \includegraphics[width=\textwidth]{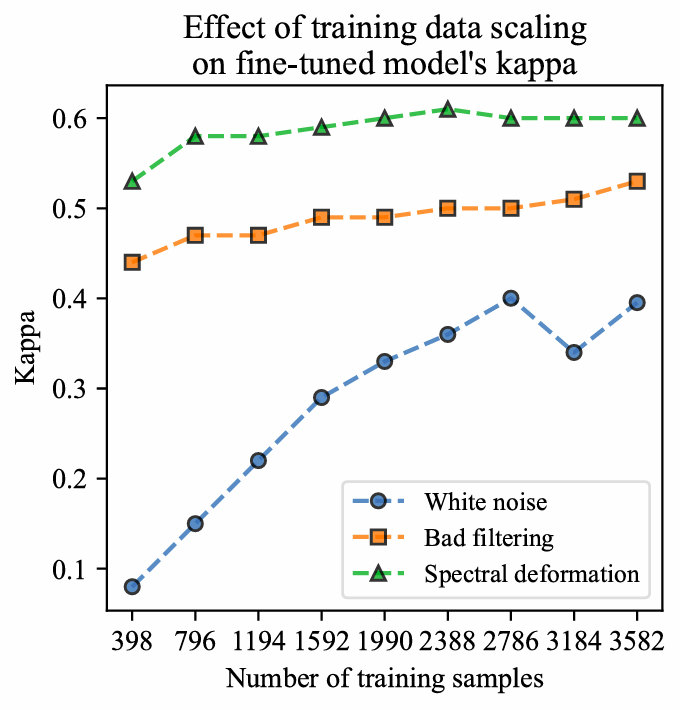}
    \end{subfigure}
\caption{Effect of the training set size on the fine-tuned model’s Kappa under different noise conditions. The plot compares Kappa values obtained with white noise and spectral deformation and amplifier overloaded distortions as the proportion of training samples increases from 10\% to 90\%.}
\label{fig:training_reports}
\end{figure}

Figure~\ref{fig:training_reports} illustrates the effect of training set size on the fine-tuned model’s performance, for each distortion type. The total number of available training nights is 3980, and the x-axis represents the number of samples used during training.

As the training set size increases, the model’s performance consistently improves across all conditions, albeit with diminishing returns.

\section{Discussion}
\label{Conclusion}

'Discriminator-guided fine  tuning', as we call this approach, was originally inspired by discussions with experienced, human sleep scorers. They mentioned that sometimes, when dealing with very problematic recordings, they would ignore certain channels or phenomena, because the resulting hypnograms would otherwise be nonsense. As researchers who are always trying to get models to better deal with new issues and domain mismatches, it was an interesting idea to design a machine learning approach mimicking the human approach. Especially since the human scorers seem to find that the approach has merit.


We find that discriminator guided fine tuning improves sleep scoring performance for almost all recordings, across distortion types. We also see that the obtained kappa values are almost always outside the 'chance' range, and the improvements are significantly better than what can be explained by chance. \textcolor{black}{However, in the present version, the fine-tuned kappa values are usually worse than what is obtained with supervised methods, and most average values are below values reported for clinical populations \cite{danker-hopfe_interrater_2004}.}
We also find that the performance depends both on which type of noise the model has to remove, but also on the specific recording - some recordings are just harder to salvage than others. We hypothesize that the distortion dependency is related to what type of changes have to be made to the model weights. In the case where one of two input channels has to be ignored, it is a simpler change of weights compared to, for instance, reshaping the power spectrum. It is possible that a different model architecture, for instance one using the spectrum as input, like SleepTransformer~\cite{phan2022sleeptransformer} would perform better for this type of distortion. Future work will have to experiment with different model designs.

\textcolor{black}{Along the same lines, it appears necessary to find ways to reduce the amount of data required. Either through reducing the model size, introducing data augmentation, or perhaps some way to better estimate the data distribution using fewer samples. As is seen in Figure \ref{fig:training_reports}, several thousand recordings are right now required to obtain good improvement. We should strive to bring this number down by at least an order of magnitude, preferably two. }

\subsection{Ablation study}
\textcolor{black}{To determine the impact of the different design choices in our algorithm, we have performed an 'ablation' study in which each choice is reversed, to see what impact it has on performance. Specifically, we have removed the 'anchor loss' term from the loss function, unfrozen the decoder, varied the number of discriminators and the batch size, changed the discriminator architecture as well as removed the positional encoding in the discriminator. We find that 2 and 3 discriminators perform the same, increasing the batch size from 1 to 2 increases performance slightly, and that all other changes decrease performance.}

\subsection{Additional findings}

We also performed various other experiments, the results of which were not interesting enough to warrant their own figures in the 'results' section, but which never the less are important to report:

(1) In the standard approach to using U-Sleep \cite{strom2024common}, multiple steps of preprocessing are carried out. In our study, the 'clean' dataset has been preproccessed according to the standard recipe (DC-removal, band-pass filtering, robust scaling and clipping), but no preprocessing is done after distortion (so, in the target domain). However, to ensure that this does not markedly impact the findings here, we carried out additional experiments where all calculations were repeated, but with preprocessing repeated after distorting the data. The only change in the results was that the 'white noise' distortion becomes a null-result; the impact on the pretrained model is decreased, and fine tuning has no effect. We assume that this is because the primary problem for the model in this case is the increased amplitude. We decided to stick with the current approach where preprocessing is not performed after data distortion, because we believe that the object under scrutiny here is the approach itself and not the performance of U-Sleep.

(2) We tested one completely realistic and challenging domain transfer: from the current source domain to the dataset described in Tabar et al 2023 \cite{tabar2023home}. Our findings were that the fine tuning had no significant effect on performance. \textcolor{black}{In investigating this negative result, we have considered multiple different explanations: inadequacy of the loss function (perhaps the anchor loss term was a mistake), differences in the type of domain mismatches observed, or simply limitations in the present version of the algorithm. While we cannot rule out that the 'distortion' equivalent to moving to the Tabar et al dataset is significantly more complex than what we have tested here, we think the most likely explanation is related to what we see in Figure \ref{fig:training_reports} - data scaling. The Tabar data set only has 120 recordings, thus making it significantly smaller than what we are working with here. This could explain the lack of a significant improvement, and we will investigate remedies for this in future work. }

\section{Conclusion}

\textcolor{black}{In conclusion, we think the approach is promising, and that the results highlight interesting properties of sleep data in general. However, it is worthwhile to investigate other implementations. In particular, it seems necessary to reduce the required amount of data before this can become a viable tool in most clinical or engineering settings.}

\section{Code}


\section{Contributions and acknowledgments}

KBM came up with the original idea and obtained funding for the project. AT and CH carried out preliminary investigations of the concept and best approaches. MA designed and implemented the final approach, and carried out all experiments and analysis. All authors read and accepted the manuscript.

Compute resources for the project were donated by the Danish e-Infrastructure Consortium, under grant DeiC-AU-N5-2024046, 'Advanced EEG Analysis'.

\printbibliography

@article{danker-hopfe_interrater_2004,
	title = {Interrater reliability between scorers from eight {European} sleep laboratories in subjects with different sleep disorders},
	volume = {13},
	issn = {1365-2869},
	url = {https://onlinelibrary.wiley.com/doi/abs/10.1046/j.1365-2869.2003.00375.x},
	doi = {10.1046/j.1365-2869.2003.00375.x},
	language = {en},
	number = {1},
	urldate = {2024-03-21},
	journal = {Journal of Sleep Research},
	author = {Danker-Hopfe, Heidi and Kunz, D. and Gruber, G. and Klösch, G. and Lorenzo, J. L. and Himanen, S. L. and Kemp, B. and Penzel, T. and Röschke, J. and Dorn, H. and Schlögl, A. and Trenker, E. and Dorffner, G.},
	year = {2004},
	note = {\_eprint: https://onlinelibrary.wiley.com/doi/pdf/10.1046/j.1365-2869.2003.00375.x},
	pages = {63--69},
}

@article{nuwer201810-10,
	title = {10-10 electrode system for {EEG} recording},
	volume = {129},
	issn = {1388-2457},
	url = {https://www.sciencedirect.com/science/article/pii/S1388245718300907},
	doi = {10.1016/j.clinph.2018.01.065},
	pages = {1103},
	number = {5},
	journaltitle = {Clinical Neurophysiology},
	shortjournal = {Clinical Neurophysiology},
	author = {Nuwer, Marc R.},
	urldate = {2024-02-27},
	date = {2018-05-01},
}

@article{gorjan2022removal,
	title = {Removal of movement-induced {EEG} artifacts: current state of the art and guidelines},
	volume = {19},
	issn = {1741-2552},
	shorttitle = {Removal of movement-induced {EEG} artifacts},
	url = {https://doi.org/10.1088/1741-2552/ac542c},
	doi = {10.1088/1741-2552/ac542c},
	language = {en},
	number = {1},
	urldate = {2026-01-15},
	journal = {Journal of Neural Engineering},
	author = {Gorjan, Dasa and Gramann, Klaus and De Pauw, Kevin and Marusic, Uros},
	month = feb,
	year = {2022},
	note = {Publisher: IOP Publishing},
	pages = {011004},
}

@article{kalita2024aneeg,
	title = {{AnEEG}: leveraging deep learning for effective artifact removal in {EEG} data},
	volume = {14},
	rights = {2024 The Author(s)},
	issn = {2045-2322},
	url = {https://www.nature.com/articles/s41598-024-75091-z},
	doi = {10.1038/s41598-024-75091-z},
	shorttitle = {{AnEEG}},
	pages = {24234},
	number = {1},
	journaltitle = {Scientific Reports},
	shortjournal = {Sci Rep},
	author = {Kalita, Bhabesh and Deb, Nabamita and Das, Daisy},
	urldate = {2026-01-15},
	date = {2024-10-16},
	langid = {english},
	note = {Publisher: Nature Publishing Group},
}

@article{raj2025comprehensive,
	title = {A comprehensive review of deep learning models for denoising {EEG} signals: challenges, advances, and future directions},
	volume = {7},
	issn = {3004-9261},
	url = {https://doi.org/10.1007/s42452-025-07808-2},
	doi = {10.1007/s42452-025-07808-2},
	shorttitle = {A comprehensive review of deep learning models for denoising {EEG} signals},
	pages = {1268},
	number = {11},
	journaltitle = {Discover Applied Sciences},
	shortjournal = {Discov Appl Sci},
	author = {Raj, Vandana Akshath and Parupudi, Tejasvi and Thalengala, Ananthakrishna and Nayak, Subramanya G.},
	urldate = {2026-01-15},
	date = {2025-10-22},
	langid = {english},
}

@article{kleeva2024resting,
	title = {Resting-state {EEG} recorded with gel-based vs. consumer dry electrodes: spectral characteristics and across-device correlations},
	volume = {18},
	issn = {1662-453X},
	url = {https://www.frontiersin.org/journals/neuroscience/articles/10.3389/fnins.2024.1326139/full},
	doi = {10.3389/fnins.2024.1326139},
	shorttitle = {Resting-state {EEG} recorded with gel-based vs. consumer dry electrodes},
	journaltitle = {Frontiers in Neuroscience},
	shortjournal = {Front. Neurosci.},
	author = {Kleeva, Daria and Ninenko, Ivan and Lebedev, Mikhail A.},
	urldate = {2026-01-14},
	date = {2024-02-02},
	note = {Publisher: Frontiers},
}

@article{leach2020protocol,
	title = {A Protocol for Comparing Dry and Wet {EEG} Electrodes During Sleep},
	volume = {14},
	issn = {1662-4548},
	doi = {10.3389/fnins.2020.00586},
	pages = {586},
	journaltitle = {Frontiers in Neuroscience},
	shortjournal = {Front Neurosci},
	author = {Leach, Sven and Chung, Ku-Young and Tüshaus, Laura and Huber, Reto and Karlen, Walter},
	date = {2020},
	pmid = {32625053},
	pmcid = {PMC7313551},
}

@article{yoo2021transferring,
  title={Transferring structured knowledge in unsupervised domain adaptation of a sleep staging network},
  author={Yoo, Chaehwa and Lee, Hyang Woon and Kang, Je-Won},
  journal={IEEE journal of biomedical and health informatics},
  volume={26},
  number={3},
  pages={1273--1284},
  year={2021},
  publisher={IEEE}
}

@article{rosen2012multisite,
	title = {A multisite randomized trial of portable sleep studies and positive airway pressure autotitration versus laboratory-based polysomnography for the diagnosis and treatment of obstructive sleep apnea: the {HomePAP} study},
	volume = {35},
	issn = {1550-9109},
	shorttitle = {A multisite randomized trial of portable sleep studies and positive airway pressure autotitration versus laboratory-based polysomnography for the diagnosis and treatment of obstructive sleep apnea},
	doi = {10.5665/sleep.1870},
	language = {eng},
	number = {6},
	journal = {Sleep},
	author = {Rosen, Carol L. and Auckley, Dennis and Benca, Ruth and Foldvary-Schaefer, Nancy and Iber, Conrad and Kapur, Vishesh and Rueschman, Michael and Zee, Phyllis and Redline, Susan},
	month = jun,
	year = {2012},
	pmid = {22654195},
	pmcid = {PMC3353048},
	pages = {757--767},
}

@article{mikkelsen2025ear,
	title = {Ear-{EEG} sleep monitoring data sets},
	volume = {12},
	issn = {2052-4463},
	url = {https://doi.org/10.1038/s41597-025-04579-8},
	doi = {10.1038/s41597-025-04579-8},
	number = {1},
	journal = {Scientific Data},
	author = {Mikkelsen, Kaare Bjarke and Rezai Tabar, Yousef and Rævsbæk Birch, Laura and Lind Kappel, Simon and Bech Christensen, Christian and Dalskov Mosgaard, Lars and Otto, Marit and Christian Hemmsen, Martin and Lind Rank, Mike and Kidmose, Preben},
	month = feb,
	year = {2025},
	pages = {301},
}

@inproceedings{lebiecka-johansen2025benefits,
	title = {Benefits of {Different} {Strategies} to {Adapt} {Sleep} {Scoring} {Models} from {Scalp}- to {Ear}-{EEG}},
	url = {https://ieeexplore.ieee.org/document/11252780},
	doi = {10.1109/EMBC58623.2025.11252780},
	urldate = {2026-01-08},
	booktitle = {2025 47th {Annual} {International} {Conference} of the {IEEE} {Engineering} in {Medicine} and {Biology} {Society} ({EMBC})},
	author = {Lebiecka-Johansen, Patrycja and Strøm, Jesper and Mikkelsen, Kaare B. and Cabrera, Alvaro F. and Madsen, Rasmus E. and Christensen, Julie A. E. and Hemmsen, Martin C. and Kidmose, Preben},
	month = jul,
	year = {2025},
	note = {ISSN: 2694-0604},
	pages = {1--7},
}

@article{fiorillo2023usleeps,
	title = {U-{Sleep}’s resilience to {AASM} guidelines},
	volume = {6},
	copyright = {2023 The Author(s)},
	issn = {2398-6352},
	url = {https://www.nature.com/articles/s41746-023-00784-0},
	doi = {10.1038/s41746-023-00784-0},
	language = {en},
	number = {1},
	urldate = {2023-08-01},
	journal = {npj Digital Medicine},
	author = {Fiorillo, Luigi and Monachino, Giuliana and van der Meer, Julia and Pesce, Marco and Warncke, Jan D. and Schmidt, Markus H. and Bassetti, Claudio L. A. and Tzovara, Athina and Favaro, Paolo and Faraci, Francesca D.},
	month = mar,
	year = {2023},
	note = {Number: 1
Publisher: Nature Publishing Group},
	pages = {1--9},
}

@article{zhang2018national,
	title = {The {National} {Sleep} {Research} {Resource}: towards a sleep data commons},
	volume = {25},
	issn = {1527-974X},
	shorttitle = {The {National} {Sleep} {Research} {Resource}},
	doi = {10.1093/jamia/ocy064},
	language = {eng},
	number = {10},
	journal = {Journal of the American Medical Informatics Association: JAMIA},
	author = {Zhang, Guo-Qiang and Cui, Licong and Mueller, Remo and Tao, Shiqiang and Kim, Matthew and Rueschman, Michael and Mariani, Sara and Mobley, Daniel and Redline, Susan},
	month = oct,
	year = {2018},
	pmid = {29860441},
	pmcid = {PMC6188513},
	pages = {1351--1358},
}

@article{bakker2018gastric,
	title = {Gastric {Banding} {Surgery} versus {Continuous} {Positive} {Airway} {Pressure} for {Obstructive} {Sleep} {Apnea}: {A} {Randomized} {Controlled} {Trial}},
	volume = {197},
	issn = {1535-4970},
	shorttitle = {Gastric {Banding} {Surgery} versus {Continuous} {Positive} {Airway} {Pressure} for {Obstructive} {Sleep} {Apnea}},
	doi = {10.1164/rccm.201708-1637LE},
	language = {eng},
	number = {8},
	journal = {American Journal of Respiratory and Critical Care Medicine},
	author = {Bakker, Jessie P. and Tavakkoli, Ali and Rueschman, Michael and Wang, Wei and Andrews, Robert and Malhotra, Atul and Owens, Robert L. and Anand, Amit and Dudley, Katherine A. and Patel, Sanjay R.},
	month = apr,
	year = {2018},
	pmid = {29035093},
	pmcid = {PMC5909166},
	pages = {1080--1083},
}

@article{oreilly2014montreal,
	title = {Montreal {Archive} of {Sleep} {Studies}: an open-access resource for instrument benchmarking and exploratory research},
	volume = {23},
	copyright = {© 2014 European Sleep Research Society},
	issn = {1365-2869},
	shorttitle = {Montreal {Archive} of {Sleep} {Studies}},
	url = {https://onlinelibrary.wiley.com/doi/abs/10.1111/jsr.12169},
	doi = {https://doi.org/10.1111/jsr.12169},
	language = {en},
	number = {6},
	urldate = {2021-03-02},
	journal = {Journal of Sleep Research},
	author = {O'Reilly, Christian and Gosselin, Nadia and Carrier, Julie and Nielsen, Tore},
	year = {2014},
	note = {\_eprint: https://onlinelibrary.wiley.com/doi/pdf/10.1111/jsr.12169},
	pages = {628--635},
}

@article{kemp2000analysis,
  title={Analysis of a sleep-dependent neuronal feedback loop: the slow-wave microcontinuity of the EEG},
  author={Kemp, Bob and Zwinderman, Aeilko H and Tuk, Bert and Kamphuisen, Hilbert AC and Oberye, Josefien JL},
  journal={IEEE Transactions on Biomedical Engineering},
  volume={47},
  number={9},
  pages={1185--1194},
  year={2000},
  publisher={IEEE}
}

@article{goldberger2000physiobank,
  title={PhysioBank, PhysioToolkit, and PhysioNet: components of a new research resource for complex physiologic signals},
  author={Goldberger, Ary L and Amaral, Luis AN and Glass, Leon and Hausdorff, Jeffrey M and Ivanov, Plamen Ch and Mark, Roger G and Mietus, Joseph E and Moody, George B and Peng, Chung-Kang and Stanley, H Eugene},
  journal={circulation},
  volume={101},
  number={23},
  pages={e215--e220},
  year={2000},
  publisher={Lippincott Williams \& Wilkins}
}

@article{spira2008sleep,
  title={Sleep-disordered breathing and cognition in older women},
  author={Spira, Adam P and Blackwell, Terri and Stone, Katie L and Redline, Susan and Cauley, Jane A and Ancoli-Israel, Sonia and Yaffe, Kristine},
  journal={Journal of the American Geriatrics Society},
  volume={56},
  number={1},
  pages={45--50},
  year={2008},
  publisher={Wiley Online Library}
}

@article{khalighi2016isruc,
  title={ISRUC-Sleep: A comprehensive public dataset for sleep researchers},
  author={Khalighi, Sirvan and Sousa, Teresa and Santos, Jos{\'e} Moutinho and Nunes, Urbano},
  journal={Computer methods and programs in biomedicine},
  volume={124},
  pages={180--192},
  year={2016},
  publisher={Elsevier}
}

@article{guillot2020dreem,
  title={Dreem open datasets: Multi-scored sleep datasets to compare human and automated sleep staging},
  author={Guillot, Antoine and Sauvet, Fabien and During, Emmanuel H and Thorey, Valentin},
  journal={IEEE transactions on neural systems and rehabilitation engineering},
  volume={28},
  number={9},
  pages={1955--1965},
  year={2020},
  publisher={IEEE}
}

@article{redline1995familial,
  title={The familial aggregation of obstructive sleep apnea.},
  author={Redline, Susan and Tishler, Peter V and Tosteson, Tor D and Williamson, John and Kump, Kenneth and Browner, Ilene and Ferrette, Veronica and Krejci, Patrick},
  journal={American journal of respiratory and critical care medicine},
  volume={151},
  number={3},
  pages={682--687},
  year={1995},
  publisher={American Public Health Association}
}

@inproceedings{ghassemi2018you,
  title={You snooze, you win: the physionet/computing in cardiology challenge 2018},
  author={Ghassemi, Mohammad M and Moody, Benjamin E and Lehman, Li-Wei H and Song, Christopher and Li, Qiao and Sun, Haoqi and Mark, Roger G and Westover, M Brandon and Clifford, Gari D},
  booktitle={2018 Computing in Cardiology Conference (CinC)},
  volume={45},
  pages={1--4},
  year={2018},
  organization={IEEE}
}

@article{phan2022sleeptransformer,
  title={Sleeptransformer: Automatic sleep staging with interpretability and uncertainty quantification},
  author={Phan, Huy and Mikkelsen, Kaare and Ch{\'e}n, Oliver Y and Koch, Philipp and Mertins, Alfred and De Vos, Maarten},
  journal={IEEE Transactions on Biomedical Engineering},
  volume={69},
  number={8},
  pages={2456--2467},
  year={2022},
  publisher={IEEE}
}

@article{tabar2023home,
  title={At-home sleep monitoring using generic ear-EEG},
  author={Tabar, Yousef R and Mikkelsen, Kaare B and Shenton, Nelly and Kappel, Simon L and Bertelsen, Astrid R and Nikbakht, Reza and Toft, Hans O and Henriksen, Chris H and Hemmsen, Martin C and Rank, Mike L and others},
  journal={Frontiers in neuroscience},
  volume={17},
  pages={987578},
  year={2023},
  publisher={Frontiers Media SA}
}

@article{heremans2022unsupervised,
  title={From unsupervised to semi-supervised adversarial domain adaptation in electroencephalography-based sleep staging},
  author={Heremans, Elisabeth RM and Phan, Huy and Borz{\'e}e, Pascal and Buyse, Bertien and Testelmans, Dries and De Vos, Maarten},
  journal={Journal of Neural Engineering},
  volume={19},
  number={3},
  pages={036044},
  year={2022},
  publisher={IOP Publishing}
}

@article{ganin2016domain,
  title={Domain-adversarial training of neural networks},
  author={Ganin, Yaroslav and Ustinova, Evgeniya and Ajakan, Hana and Germain, Pascal and Larochelle, Hugo and Laviolette, Fran{\c{c}}ois and March, Mario and Lempitsky, Victor},
  journal={Journal of machine learning research},
  volume={17},
  number={59},
  pages={1--35},
  year={2016}
}

@article{kiasari2018joint,
  title={Joint moment-matching autoencoders},
  author={Kiasari, Mohammad Ahangar and Moirangthem, Dennis Singh and Lee, Minho},
  journal={Neural Networks},
  volume={106},
  pages={185--193},
  year={2018},
  publisher={Elsevier}
}

@inproceedings{hsu2017unsupervised,
  title={Unsupervised domain adaptation for robust speech recognition via variational autoencoder-based data augmentation},
  author={Hsu, Wei-Ning and Zhang, Yu and Glass, James},
  booktitle={2017 IEEE automatic speech recognition and understanding workshop (ASRU)},
  pages={16--23},
  year={2017},
  organization={IEEE}
}

@collection{altevogt2006sleep,
	location = {Washington ({DC})},
	title = {Sleep Disorders and Sleep Deprivation: An Unmet Public Health Problem},
	rights = {Copyright © 2006, National Academy of Sciences.},
	isbn = {978-0-309-10111-0},
	url = {http://www.ncbi.nlm.nih.gov/books/NBK19960/},
	series = {The National Academies Collection: Reports funded by National Institutes of Health},
	shorttitle = {Sleep Disorders and Sleep Deprivation},
	urldate = {2026-04-06},
	date = {2006},
	pmid = {20669438},
}

@article{sateia_international_2014,
	title = {International classification of sleep disorders-third edition: highlights and modifications},
	volume = {146},
	issn = {1931-3543},
	doi = {10.1378/chest.14-0970},
	shorttitle = {International classification of sleep disorders-third edition},
	pages = {1387--1394},
	number = {5},
	journaltitle = {Chest},
	shortjournal = {Chest},
	author = {Sateia, Michael J.},
	date = {2014-11},
	pmid = {25367475},
}

@article{phan2022automatic,
  title={Automatic sleep staging of EEG signals: recent development, challenges, and future directions},
  author={Phan, Huy and Mikkelsen, Kaare},
  journal={Physiological Measurement},
  volume={43},
  number={4},
  pages={04TR01},
  year={2022},
  publisher={IOP Publishing}
}

@article{american2007aasm,
  title={The AASM manual for the scoring of sleep and associated events: rules, terminology and technical specifications},
  author={American Academy of Sleep Medicine and others},
  journal={Westchester, IL: American Academy of Sleep Medicine},
  volume={23},
  year={2007}
}

@article{strom2024common,
  title={Common sleep data pipeline for combined data sets},
  author={Str{\o}m, Jesper and Engholm, Andreas Larsen and Lorenzen, Kristian Peter and Mikkelsen, Kaare B},
  journal={Plos one},
  volume={19},
  number={8},
  pages={e0307202},
  year={2024},
  publisher={Public Library of Science San Francisco, CA USA}
}

@article{fiorillo2019automated,
  title={Automated sleep scoring: A review of the latest approaches},
  author={Fiorillo, Luigi and Puiatti, Alessandro and Papandrea, Michela and Ratti, Pietro-Luca and Favaro, Paolo and Roth, Corinne and Bargiotas, Panagiotis and Bassetti, Claudio L and Faraci, Francesca D},
  journal={Sleep medicine reviews},
  volume={48},
  pages={101204},
  year={2019},
  publisher={Elsevier}
}

@article{perslev2021u,
  title={U-Sleep: resilient high-frequency sleep staging},
  author={Perslev, Mathias and Darkner, Sune and Kempfner, Lykke and Nikolic, Miki and Jennum, Poul J{\o}rgen and Igel, Christian},
  journal={NPJ digital medicine},
  volume={4},
  number={1},
  pages={72},
  year={2021},
  publisher={Nature Publishing Group UK London}
}

@article{mikkelsen2019accurate,
  title={Accurate whole-night sleep monitoring with dry-contact ear-EEG},
  author={Mikkelsen, Kaare B and Tabar, Yousef R and Kappel, Simon L and Christensen, Christian B and Toft, Hans O and Hemmsen, Martin C and Rank, Mike L and Otto, Marit and Kidmose, Preben},
  journal={Scientific reports},
  volume={9},
  number={1},
  pages={16824},
  year={2019},
  publisher={Nature Publishing Group UK London}
}

@book{olivas2009handbook,
  title={Handbook of research on machine learning applications and trends: Algorithms, methods, and techniques: Algorithms, methods, and techniques},
  author={Olivas, Emilio Soria and Guerrero, Jos David Mart and Martinez-Sober, Marcelino and Magdalena-Benedito, Jose Rafael and Serrano, L and others},
  year={2009},
  publisher={IGI global}
}

@article{fan2022unsupervised,
  title={Unsupervised domain adaptation by statistics alignment for deep sleep staging networks},
  author={Fan, Jiahao and Zhu, Hangyu and Jiang, Xinyu and Meng, Long and Chen, Chen and Fu, Cong and Yu, Huan and Dai, Chenyun and Chen, Wei},
  journal={IEEE Transactions on Neural Systems and Rehabilitation Engineering},
  volume={30},
  pages={205--216},
  year={2022},
  publisher={IEEE}
}

@article{lorenzen2024personalization,
  title={Personalization of automatic sleep scoring: How best to adapt models to personal domains in wearable EEG},
  author={Lorenzen, Kristian P and Heremans, Elisabeth RM and de Vos, Maarten and Mikkelsen, Kaare B},
  journal={IEEE Journal of Biomedical and Health Informatics},
  year={2024},
  publisher={IEEE}
}

@article{cohen1960coefficient,
  title={A coefficient of agreement for nominal scales},
  author={Cohen, Jacob},
  journal={Educational and psychological measurement},
  volume={20},
  number={1},
  pages={37--46},
  year={1960},
  publisher={Sage Publications Sage CA: Thousand Oaks, CA}
}

@article{phan2020personalized,
  title={Personalized automatic sleep staging with single-night data: a pilot study with Kullback--Leibler divergence regularization},
  author={Phan, Huy and Mikkelsen, Kaare and Ch{\'e}n, Oliver Y and Koch, Philipp and Mertins, Alfred and Kidmose, Preben and De Vos, Maarten},
  journal={Physiological measurement},
  volume={41},
  number={6},
  pages={064004},
  year={2020},
  publisher={IOP Publishing}
}



\newpage
\newpage

\FloatBarrier
\appendix

\section{Appendix}
\subsection{Distortion details}

\begin{figure*}[htb]
    \centering
    \begin{subfigure}[t]{1\textwidth}
        \centering
        \includegraphics[width=\textwidth]{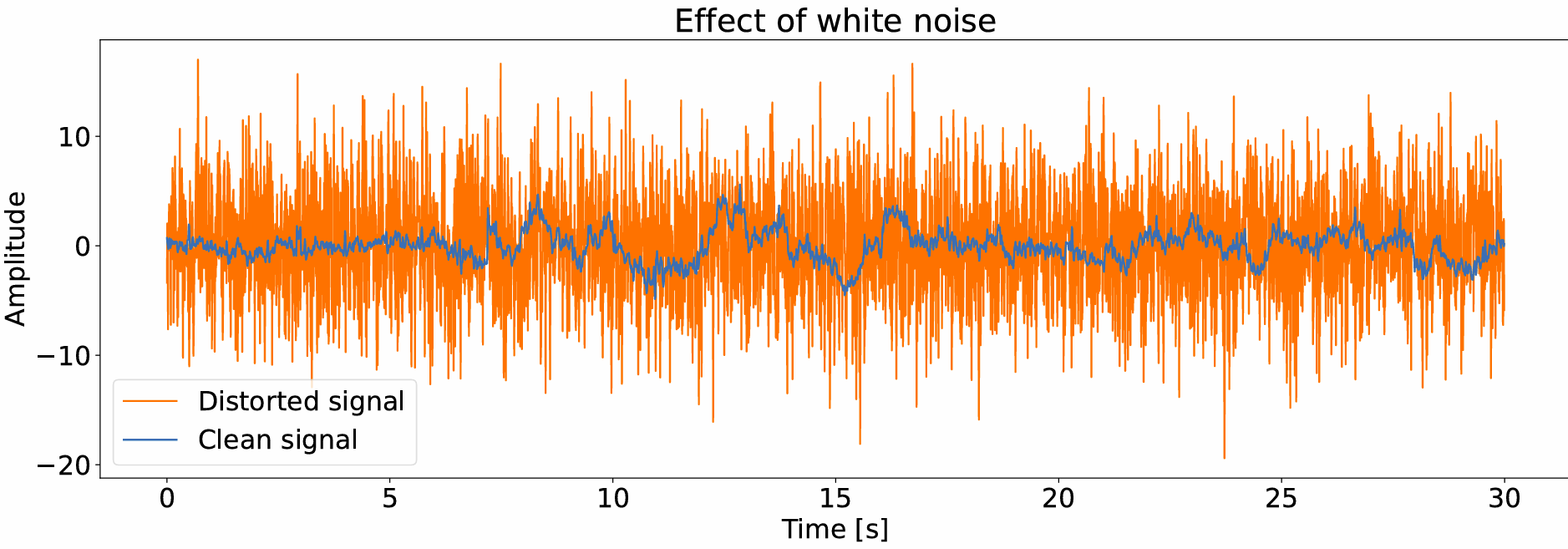}
        \caption{White noise deformation}
    \end{subfigure}
    \begin{subfigure}[t]{1\textwidth}
        \centering
        \includegraphics[width=\textwidth]{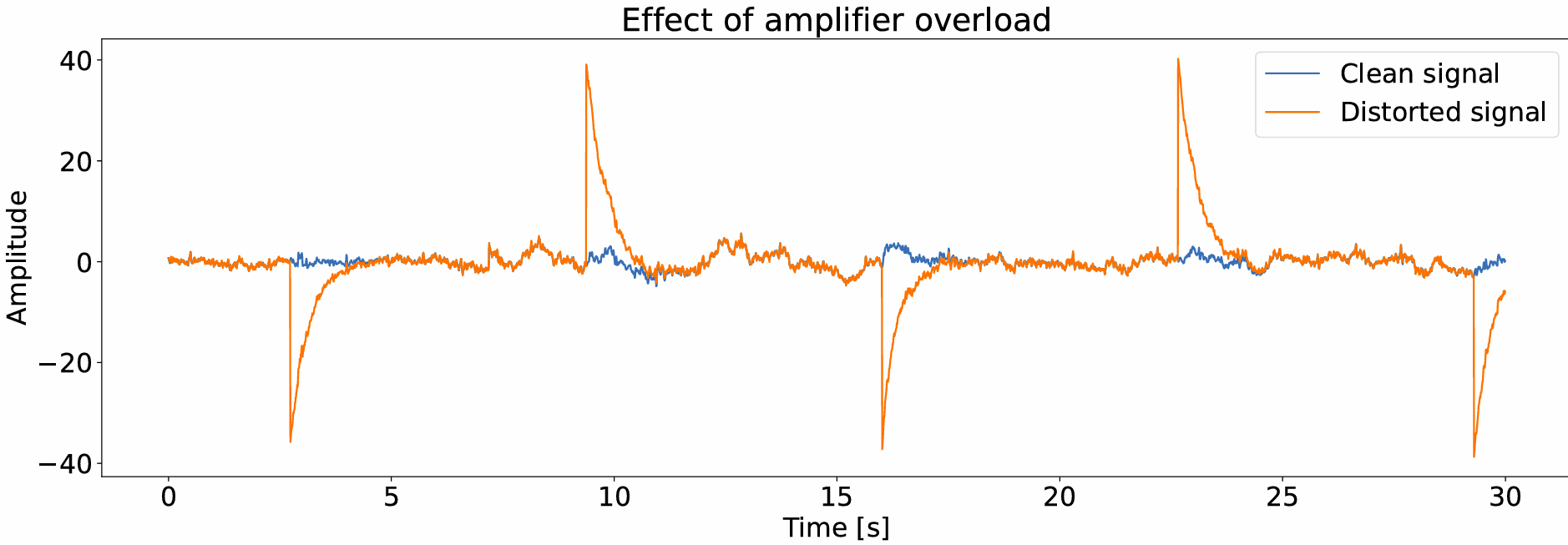}
        \caption{Amplifier overload}
    \end{subfigure}
    \begin{subfigure}[t]{1\textwidth}
        \centering
        \includegraphics[width=\textwidth]{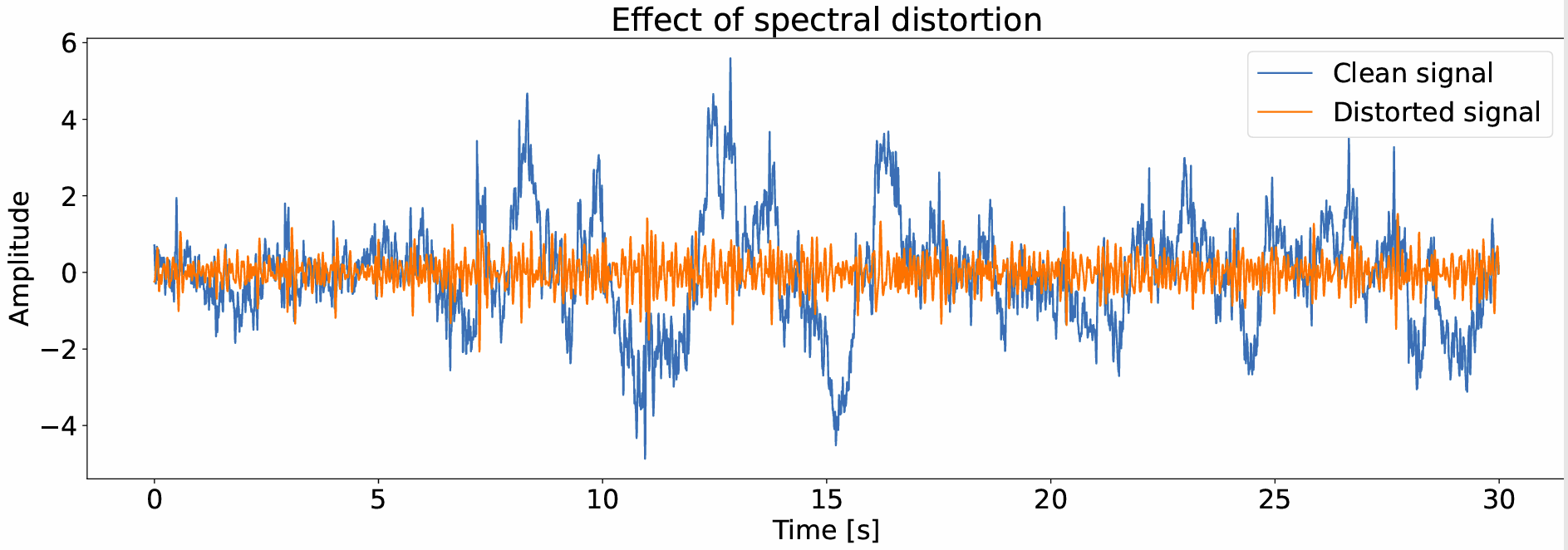}
        \caption{Spectral distortion.}
    \end{subfigure}
    \caption{(a)--(c) Visualizations of three representative distortions: bad filtering, amplifier overload, and additive white noise. The blue curve denotes the clean signal, and the red (dotted) curve denotes the distorted signal.}
    \label{fig:three_distortion_visualize}
\end{figure*}

\textcolor{black}{
Fig \ref{fig:three_distortion_visualize} compares the original waveform with the three distortion types; spectral deformation, amplifier overload, and additive white noise, highlighting how each distortion alters the temporal structure in a distinct manner.}

\begin{figure}[ht]
    \centering
    \begin{subfigure}[t]{0.48\textwidth}
        \centering
        \includegraphics[width=\textwidth]{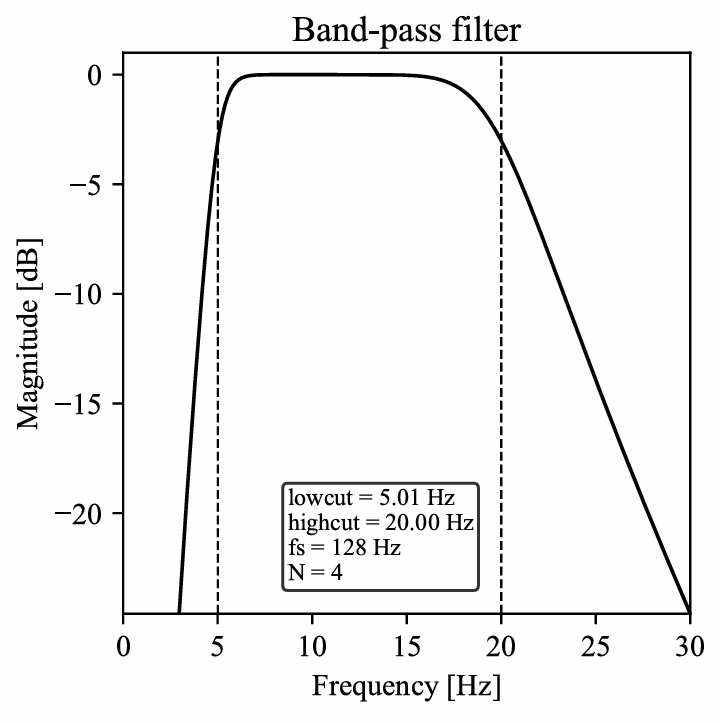}
    \end{subfigure}
\caption{An example of band-pass filter applied to the signal with a low cut-off of 5.00\,Hz and a high cut-off of 20.0\,Hz, suppressing low-frequency drift and high-frequency noise while retaining features in the 5–20\,Hz band.}
\label{fig:plot_bad_filter}
\end{figure}

\textcolor{black}{Additionally, in Figure \ref{fig:plot_bad_filter} is shown the amplitude characteristic of the type of bandpass filter used to create 'spectral deformation'. }

\FloatBarrier

\subsection{Algorithm for calculating chance kappa distribution}

\begin{algorithmic}[1]
\REQUIRE Fine-tuned U-Sleep model $U_f$, target domain consisting of distorted sleep recording $ \mathcal{D}_t$. Let N be number of experiments, and $\mathcal{K}$ the Chance Kappa distribution.\\
$\mathcal{K} \gets [\,]$
\FOR{each i in  N}
    \STATE Randomly sample $x_i$ and $y_i$ from $\mathcal{D}_t$
    \STATE Predict the hypnogram:
    $h_{x_i} \gets U_f(x_i)$ ,
    $h_{y_i} \gets U_f(y_i)$
    \STATE Compute kappa score between hypnograms $h_{y_i}, h_{x_i}$ : $k_i \gets \kappa (h_{y_i}, h_{x_i})$
    \STATE Appending : $\mathcal{K} \gets \{k_i\}$
\ENDFOR \\

Analysis distribution of $\mathcal{K}$\;
\end{algorithmic}
\FloatBarrier

\subsection{Training curves}

\begin{figure*}[h]
    \centering
    \begin{subfigure}[t]{0.4\textwidth}
        \centering
        \includegraphics[width=\textwidth]{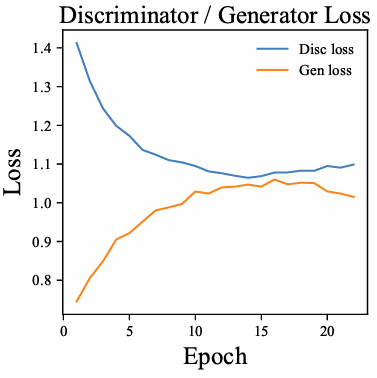}
        \caption{}
    \end{subfigure}
    \begin{subfigure}[t]{0.4\textwidth}
        \centering
        \includegraphics[width=\textwidth]{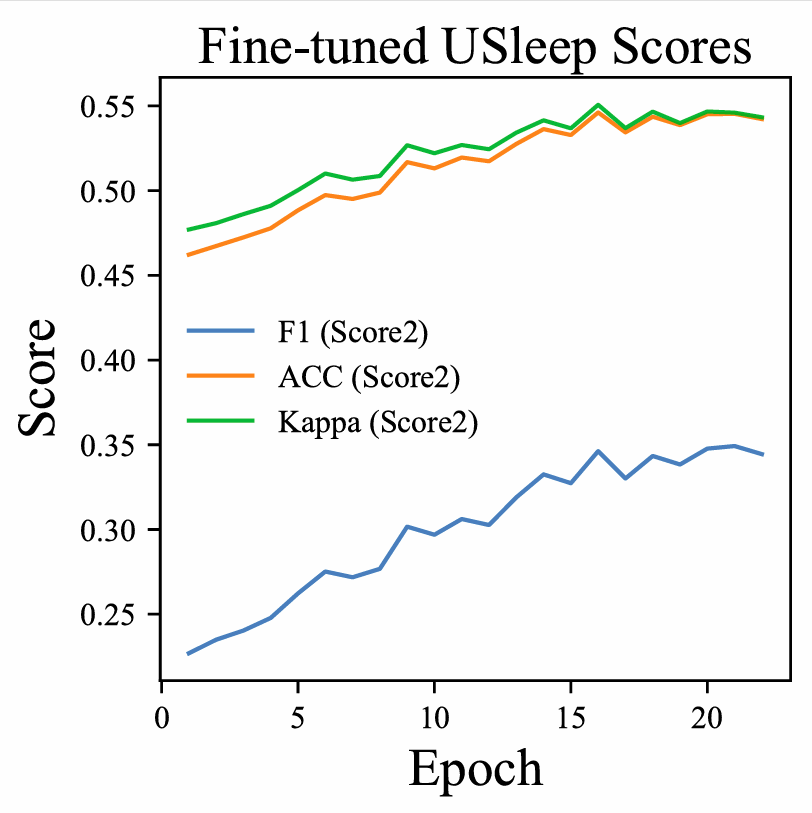}
        \caption{}
    \end{subfigure}
\caption{(a) Training curves of the generator and discriminator. (b) Curves of F1-score, accuracy, and Cohen’s kappa on the test set during training, for the 'spectral deformation' distortion. }
\label{fig:loss_curves_appendix}
\end{figure*}

\textcolor{black}{In Figure \ref{fig:loss_curves_appendix} is shown both the losses for the generator and the discriminators, as well as a performance scores (F1, accuracy, kappa), for a single training run, using 'spectral deformation' as the distortion type. We see that, as expected, the discriminator and generator losses are negatively correlated, but also that, interestingly, the performance measures do not decrease as soon as the generator loss. This is positive because it means that we are not quite as sensitive to the precise stopping time as we might otherwise have been. }

\FloatBarrier

\subsection{Performance scatterplot}

\begin{figure*}[htb]
    \centering
    \begin{subfigure}[t]{0.4\textwidth}
        \centering
        \includegraphics[width=\textwidth]{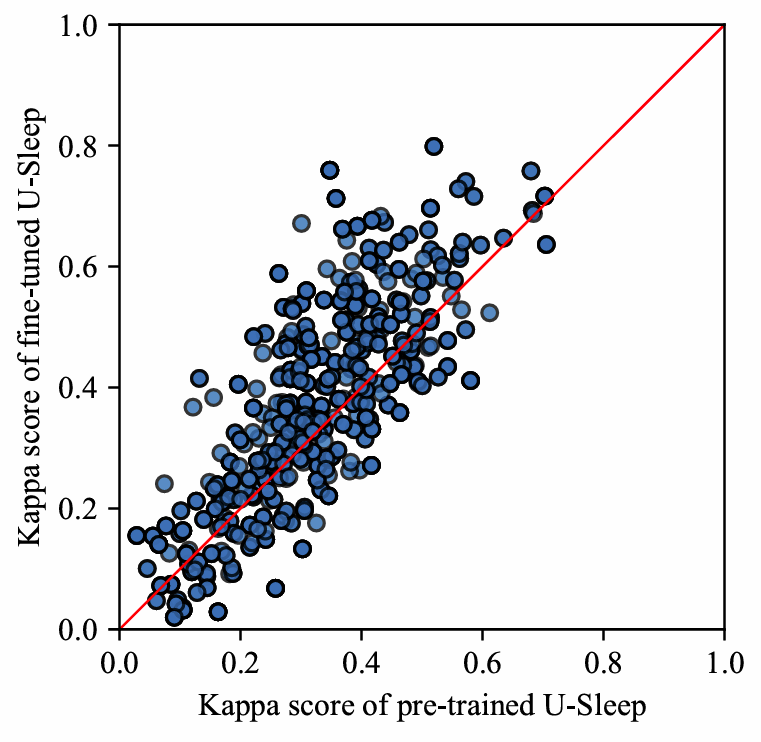}
        \caption{}
    \end{subfigure}
    \begin{subfigure}[t]{0.4\textwidth}
        \centering
        \includegraphics[width=\textwidth]{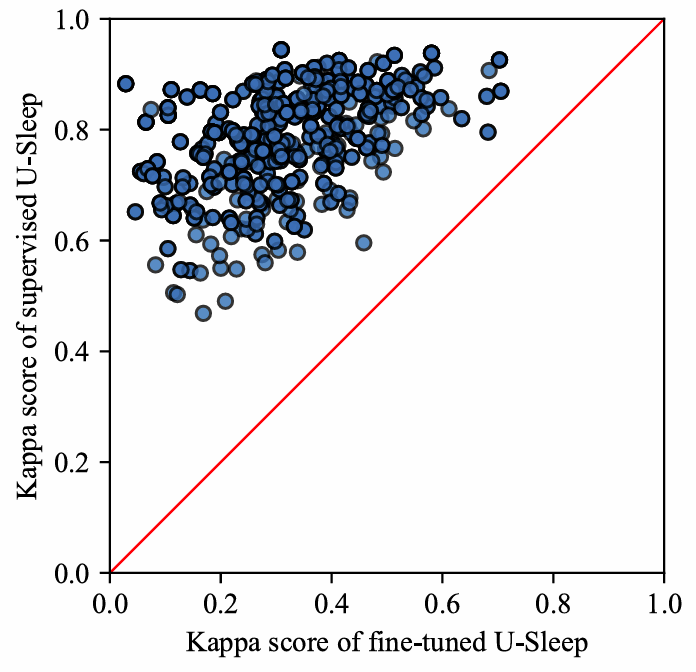}
        \caption{}
    \end{subfigure}

\caption{Comparison of pairwise kappa scores among the pre-trained, fine-tuned, and supervisedly trained U-Sleep models. The diagonal red line represents the line of equality ($y = x$).  (a) illustrates the comparison between the pre-trained and fine-tuned U-Sleep models, and (b) shows the comparison between the fine-tuned and supervisedly trained U-Sleep models.}
\label{fig:kappa_score_fine_supervised}
\end{figure*}

In Figure~\ref{fig:kappa_score_fine_supervised} upward deviations from the red line show that the benchmark model generally achieves higher agreement.

\end{document}